\documentclass[journal]{IEEEtran}
\usepackage{cite}
\usepackage{graphicx}
\usepackage{url}
\usepackage{amsmath}
\usepackage{amssymb}
\usepackage[caption = false, font=footnotesize]{subfig}
\usepackage{float}
\usepackage{booktabs} 
\usepackage[linkcolor=black]{hyperref}
\usepackage{multirow}
\usepackage{array}
\usepackage{color}
\usepackage{tabularx}
\usepackage{longtable}
\usepackage{tabu}
\usepackage{pifont}
\usepackage{algorithm, algorithmic}
\newcolumntype{L}[1]{>{\raggedright\let\newline\\\arraybackslash\hspace{0pt}}m{#1}}
\newcolumntype{C}[1]{>{\centering\let\newline\\\arraybackslash\hspace{0pt}}m{#1}}
\newcolumntype{R}[1]{>{\raggedleft\let\newline\\\arraybackslash\hspace{0pt}}m{#1}}

\usepackage{makecell}

\ifCLASSINFOpdf
\else
\fi
\newcommand{\bb}[1]{{\color{black}#1}}

\newcommand{\bl}[1]{{\color{black}#1}}

\newcommand{\ly}[1]{{\color{black}#1}}
\begin{document}

\title{Gap-closing Matters: Perceptual Quality Evaluation and  Optimization of Low-Light Image Enhancement}

\author{{Baoliang~Chen*}\thanks{*The authors contributed equally to this work.},
        Lingyu~Zhu*,
        Hanwei~Zhu,
        Wenhan~Yang,~\textit{Member,~IEEE},
        Linqi~Song,~\textit{Senior Member,~IEEE},
        and Shiqi~Wang,~\textit{Senior Member,~IEEE}~

\thanks{
Baoliang Chen,  Lingyu Zhu, Hanwei Zhu, and Linqi Song are with the Department of Computer Science, City University of Hong Kong, Hong Kong, China. (e-mail: \{blchen6-c, lingyzhu-c, hanwei.zhu\}@my.cityu.edu.hk, linqi.song@cityu.edu.hk). \\
Shiqi Wang is with the Department of Computer Science, City University of Hong Kong, Hong Kong, China, and also with the Shenzhen Research Institute, City University of Hong Kong, Shenzhen, China (e-mail: shiqwang@cityu.edu.hk). \\
Wenhan Yang is with Peng Cheng Laboratory, Shenzhen, China. (e-mail: yangwh@pcl.ac.cn)\\
Corresponding author:~\textit{Shiqi Wang.}
}
}

\markboth{Submitted to Transactions on Multimedia}%
{Shell \MakeLowercase{\textit{et al.}}: Bare Demo of IEEEtran.cls for Journals}

\maketitle

\begin{abstract}
\ly{
There is a growing consensus in the research community that the optimization of low-light image enhancement approaches should be guided by the visual quality perceived by end users. Despite the substantial efforts invested in the design of low-light enhancement algorithms, there has been comparatively limited focus on assessing subjective and objective quality systematically. To mitigate this gap and provide a clear path towards optimizing low-light image enhancement for better visual quality, we propose a gap-closing framework. In particular, our gap-closing framework starts with the creation of a large-scale dataset for Subjective QUality Assessment of REconstructed LOw-Light Images (SQUARE-LOL). This database serves as the foundation for studying the quality of enhanced images and conducting a comprehensive subjective user study. Subsequently, we propose an objective quality assessment measure that plays a critical role in bridging the gap between visual quality and enhancement. Finally, we demonstrate that our proposed objective quality measure can be incorporated into the process of optimizing the learning of the enhancement model toward perceptual optimality. We validate the effectiveness of our proposed framework through both the accuracy of quality prediction and the perceptual quality of image enhancement. Our database and codes are publicly available at \url{https://github.com/Baoliang93/IACA_For_Lowlight_IQA}.
}
\end{abstract}


\IEEEpeerreviewmaketitle

\section{Introduction}\label{sec:intro}
%

\IEEEPARstart{T}{he} prevalence of image acquisition devices has led to an increased demand for low-light image enhancement techniques that can improve the quality of photographs taken under inadequate lighting conditions. 
\ly{As a result of unbalanced lighting, low-light enhancement algorithms aim to mitigate the effects on visibility, color distortion, and excessive noise caused by unbalanced lighting. 
Early efforts focused on reconstructing images taken in low-light conditions to resemble those taken under normal illumination, including approaches based on Histogram Equalization (HE)~\cite{abdullah2007dynamic,coltuc2006exact,stark2000adaptive, arici2009histogram} and Retinex theory~\cite{fu2016weighted, guo2016lime, jobson1997multiscale, wang2013naturalness}. 
However, simply brightening the image without incorporating additional image priors can result in amplified noise. 
Recently, the deep learning-based methods~\cite{lore2017llnet, ren2019low, wang2019underexposed} have achieved significant breakthroughs. 
These approaches utilize extensive databases containing pairs of low and high-quality images to train enhancement mappings in a fully-supervised fashion.}

\ly{
Optimizing fully-supervised low-light enhancement methods can be challenging due to the lack of guidance from perceptually-calibrated quality models. 
Although it has been widely accepted that incorporating advanced quality measures in optimization might lead to more visually pleasant results~\cite{zhang2018unreasonable, johnson2016perceptual}, proven quality measures that behave consistently with the human perception are still largely lacking. 
Consequently, the quality model may not consistently correlate with the perceptual quality of the enhanced images, leading to unanticipated distortions such as biased color and residual noise.}
To address this issue, researchers have attempted to learn low-light enhancement models without paired supervision \cite{jiang2021enlightengan, ni2020unpaired}.
This approach involves training the model using only low or normal-light images without requiring paired data. 
To enable this, Generative Adversarial Network (GAN) \cite{goodfellow2014generative} is used to facilitate style transfer from low-light images to natural lighting conditions.
\ly{Despite the potential of these methods to utilize the scene statistics of natural images, it remains uncertain whether the learned statistics accurately reflect human perception in the absence of explicit guidance.}
Moreover, without the paired supervision, it is very challenging to recover the fine details and remove the sensor noise due to the weak supervision.


\begin{table*}[]
\centering
  \caption{Comparisons of the IQA databases.} 

\begin{tabular}{cccccc} 
\toprule
Database & { \makecell[c]{ $\#$ of Ref. \\ Images/Patches}}  & {Quality Degradation}  & { \makecell[c]{$\#$ of  Distortion \\ Types}}  &  \makecell[c]{$\#$ of Images \\ Patches } & {Testing Methodology}          
\\\midrule
LIVE \cite{sheikh2006statistical}   & 29               & {Synthetic distortions}                     & 5                & 797               & DMOS                      \\
CSIQ \cite{larson2010most}   & 30             & {Synthetic distortions}                      & 29               & 866               & DMOS                      \\
TID2013 \cite{ponomarenko2015image} & 25             & {Synthetic distortions}                      & 25               & 3,000              & MOS                       \\
BAPPS \cite{zhang2018unreasonable}  & 187.7k           & {Synthetic distortions + CNN alg. outputs}   & 425              &  375.4k                 & 2 Alternative Choices      \\
PieAPP \cite{prashnani2018pieapp}  & 200            & {Synthetic distortions + CNN alg. outputs} & 75               &  20,280                 & Probability of Preference \\
EHNQ \cite{yang2023ehnq}           & 100            & alg. outputs (traditional)                            & 15              &  1,500                 & MOS \\\midrule
Ours     & 290           &  {\makecell[c]{Low-light enhancement alg. outputs \\(traditional and CNN)} }              & 10               & 2,900              & 2 Alternative Choices     \\\bottomrule
\end{tabular}
 \label{tab:databse}
\end{table*}

Therefore, despite the demonstrated success, there are still some fundamental issues in low light enhancement regarding the algorithm evaluation and optimization for human vision. 
\ly{Specifically, the comparison, evaluation, and optimization of the perceptual quality of enhanced images become crucial. 
Though Chen \textit{et al.}  \cite{chen2022loop} have attempted to optimize the enhancement method by an objective quality model with a loop game manner, it suffers from unreliable enhancement and difficulty in optimization due to the unreliable pseudo labels. 
In this paper, we address this issue by bridging the gap between quality assessment and enhancement sequentially. 
In particular, we contribute a novel dedicated database with reliable subjective ratings of diverse images and various low-light image enhancement models, highlighting the potential challenges faced by competing quality assessment methods in evaluating the quality of enhanced images.}
Subsequently, we take a further step to develop a learning-based quality assessment measure in an effort to  assess the perceptual visual quality of the enhanced images from a variety of enhancement models automatically. 
Finally, we optimize the proven quality measure that achieves a high correlation with human perception on the baseline low-light enhancement models.
\ly{Comprehensive experimental results indicate that the proposed framework effectively bridges the gap, resulting in a significant improvement in quality assessment accuracy and ensuring superior quality of the enhanced images.}
The primary contributions of this paper can be summarized as follows,
\bb{
\begin{itemize}
\item We establish a large-scale database named SQUARE-LOL (Subjective Quality Assessment of Reconstructed Low-Light Image), which consists of diverse image contents and different enhancement methods. The database contains 2,900 images and pairwise subjective testing labels for the 13,050 pairs. It provides a solid foundation and a reliable way for gauging and optimizing the visual quality of enhanced images acquired under low-light conditions.

\item We develop a Blind Image Quality Assessment (BIQA) model, specifically designed for low-light image enhancement applications. The proposed Illumination Aware and Content Adaptive (IACA) quality assessment model is grounded on the exploitation of deep feature representations that govern the visual quality, delivering state-of-the-art performance on the proposed database.

\item We propose an optimization approach for image enhancement tasks based on the IACA model, which bridges the gap between quality evaluation and optimization in a principled way. By embedding the proposed BIQA measure, the proposed scheme outperforms the vanilla version in terms of visual quality, further demonstrating the feasibility and effectiveness of this framework.
\end{itemize}
}

\section{Related Works}

\subsection{Image Quality Assessment Databases} 
\ly{
To evaluate IQA methods, databases with human perception scores need to be constructed. Some widely used IQA databases include LIVE \cite{sheikh2006statistical}, TID2013 \cite{ponomarenko2015image}, CSIQ \cite{larson2010most}, and Waterloo Exploration Database \cite{ma2016waterloo}. These databases have been created by simulating multiple distortions to varying degrees. However, synthetic distorted images cannot accurately reflect the realistic distortions that occur during image acquisition, processing, or storage.
Besides, there has been a growing interest in quality evaluation for images in the wild, and several typical image databases have been created. These include LIVE Challenge \cite{ghadiyaram2015massive}, KonIQ-10k \cite{hosu2020koniq}, and SPAQ \cite{fang2020perceptual}. Additionally, the proliferation of AI-driven optimization algorithms, such as super-resolution, deblurring, denoising, and compression, has introduced complex artifacts and opened new avenues for image quality assessment. Correspondingly, related databases have been constructed to investigate perceptual similarities~\cite{zhang2018unreasonable,prashnani2018pieapp}. Notably, Perceptual Image-Error Assessment through Pairwise Preference (PieAPP)~\cite{prashnani2018pieapp} collects user judgments in a pairwise manner for perceptual preferences, thus providing more reliable scores.

In addition to the above-mentioned databases, other databases have been created to assess the quality of images under specific conditions. For example, in \cite{chen2014quality}, the quality assessment of enhanced images from hazy, underwater, and low-light scenarios was conducted, introducing different types of corruption. The enhanced night-time image quality dataset \cite{yang2023ehnq} contains a relatively smaller number of enhanced images.
Table \ref{tab:databse} presents a detailed comparison of databases that rely on synthetic and Convolutional Neural Networks (CNN) to generate images.

}

\subsection{Blind Image Quality Assessment Models} 
\ly{
In order to evaluate the degradation of naturalness in images, BIQA methods rely on natural scene statistics (NSS). 
A number of NSS features, such as local spatial normalized luminance coefficients, have been extensively investigated by researchers such as Mittal \textit{et al.} \cite{mittal2012no}. 
The free energy theory has also been explored for image quality assessment \cite{gu2014using,liu2017reduced}. 
To extract more discriminative features, researchers have utilized techniques like constructing pseudo-reference images \cite{min2017blind} and singular value decomposition (SVD) \cite{sadou2019blind}. 
Gu \textit{et al.} \cite{gu2016blind} applied statistical naturalness, information entropy, and structural preservation to estimate the quality of tone-mapping images.
In recent years, deep learning-driven BIQA techniques have exhibited better performance when compared to conventional approaches.
Zhang \textit{et al.} \cite{zhang2018blind} utilized the deep bilinear model to build a robust BIQA, achieving promising results for quality assessment of synthetic and realistic distortions. 
Talebi \textit{et al.} developed the Neural IMage Assessment (NIMA) model \cite{talebi2018nima}, which predicts the distribution of quality ratings on the AVA database  \cite{murray2012ava} . 
Su \textit{et al.} \cite{su2020blindly} aggregated the proposed local and global feature representation to learn a content-aware module, and an adaptive multi-scale hyper-network was subsequently proposed to predict image quality score. 
%
%
Zhu \textit{et al.} \cite{zhu2020metaiqa} proposed a novel deep meta-learning-based IQA model that can be easily adapted to unknown distortions by leveraging meta-knowledge transferred by humans. 
Bosse \textit{et al.} \cite{bosse2017deep} presented an end-to-end method that can simultaneously handle full-reference and no-reference IQA tasks.}

\begin{figure*}[]
\includegraphics[scale=0.13]{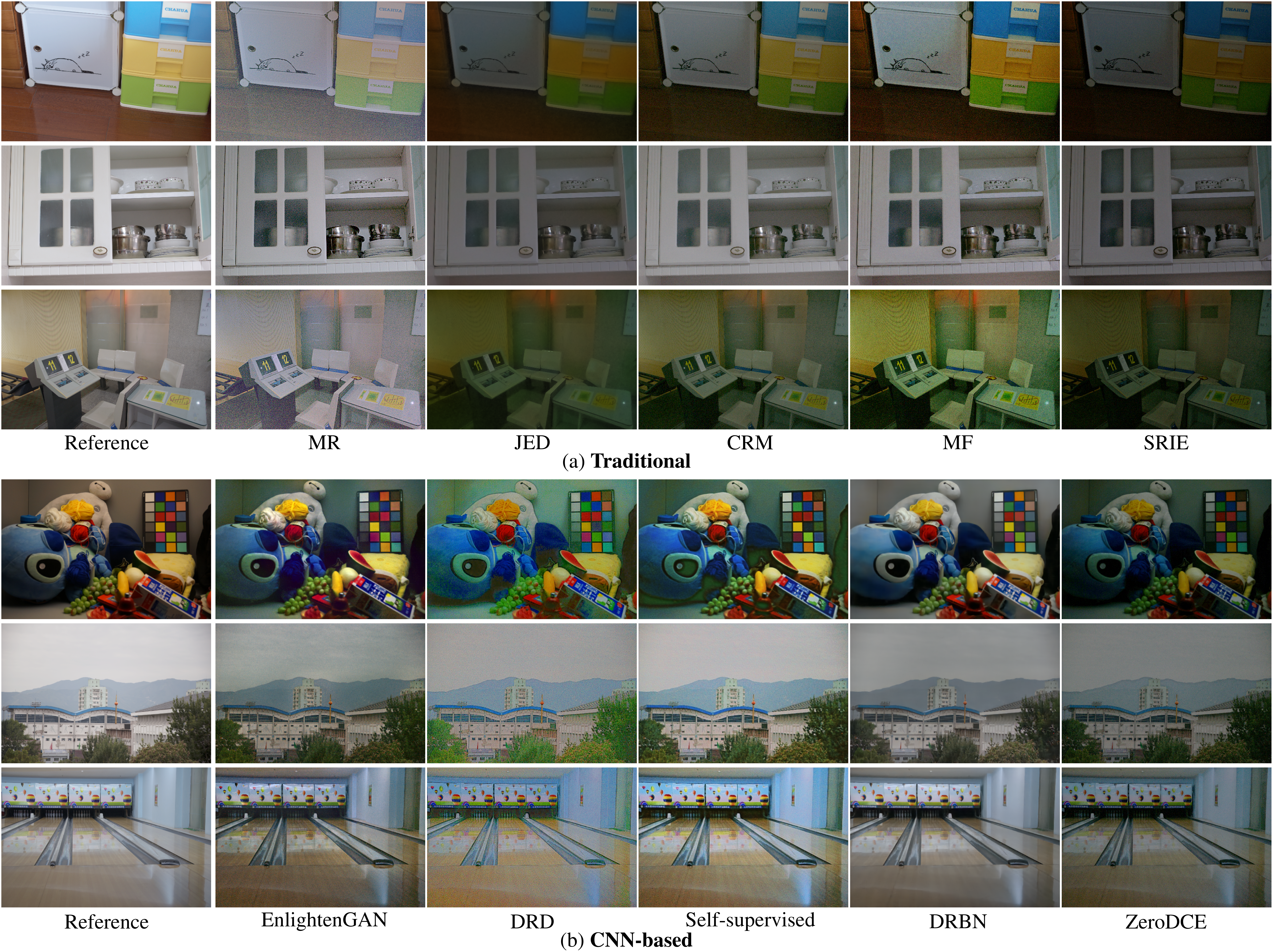}
\caption{Sampled enhanced low-light images in SQUARE-LOL database. The corresponding reference image of each scene is also provided in the first column. The images can be zoomed in to a higher magnification for a closer examination.}
\label{fig: Distortion Samples}
\end{figure*}

\subsection{Low-light Image Enhancement Methods}
\ly{
Low-light image enhancement has been extensively studied to improve the visibility of hidden information. 
Some pioneering works were developed based on physical assumptions and priors, such as Histogram Equalization (HE) \cite{abdullah2007dynamic, coltuc2006exact, stark2000adaptive, arici2009histogram} and Retinex-based methods \cite{jobson1997multiscale, jobson1997properties, land1977retinex}. 
HE is a simple and effective method that stretches the valid luminance range of given low-light images. Differential histogram equalization (DHE) \cite{nakai2013color} was a variational method that improves visual quality. 
However, HE lacks consideration of spatial details, resulting in undesirable illumination and noise. 
To overcome these limitations, Retinex theory \cite{land1964retinex} was introduced, which decomposes images into an illumination layer and reflectance layer. 
The typical Retinex-based methods include the single-scale Retinex model (SSR) \cite{jobson1997properties},  Multiscale Retinex (MSR) \cite{jobson1997multiscale}, and variational methods \cite{lee2013adaptive}. 
Fu \textit{et al.} \cite{fu2016fusion} fused the adjusted illumination and reflectance components to generate enhanced images, while Guo \textit{et al.} \cite{guo2016lime} adopted an initial illumination map by applying structure-aware prior and refining it based on an Augmented Lagrangian Multiplier (ALM) algorithm. 
Although traditional methods can improve image brightness to some extent, they still have limitations, especially when dealing with images with complex backgrounds and large noise.
In recent years, deep learning-based enhancement methods have gained increasing attention. 
For example, Low-light Net (LLNet) \cite{lore2017llnet} attempted to achieve contrast enhancement and denoising simultaneously. 
The deep recursive band network (DRBN) proposed by Yang \textit{et al.} \cite{yang2020fidelity} combined fidelity learning and adversarial learning. 
Hao \textit{et al.} \cite{hao2020low} proposed a novel semi-decoupled Retinex-based enhancement method.
The previous works \cite{li2020spectrum,wan2022purifying} introduced Near-Infrared enlightened (NIRE) images as guidance to purify low-light images.
Additionally, obtaining paired data consisting of low-light and normal-light images is often challenging. 
To overcome this limitation, EnlightenGAN, as proposed by Jiang \textit{et al.} \cite{jiang2021enlightengan}, was designed based on the principles of adversarial learning to achieve optimal perceptual quality, and Guo \textit{et al.} \cite{guo2020zero} investigated the enhancement curve and proposed zero-reference optimization for image enhancement.
}

\section{Database Construction}
\ly{
To better understand the perceptual characteristics of enhanced images and gain insights into the quality assessment of low-light image enhancement methods, we have created a dedicated database named SQUARE-LOL.
This database consists of 2,900 enhanced images, each with corresponding subjective quality ratings.
The number of source low-level images is 290, and each image is enhanced by ten classical or state-of-the-art low-level image enhancement methods, including traditional methods and deep learning-based methods.
This large-scale database allows us to reveal that a well-designed IQA model and readily-deployable optimization framework can be developed to improve the performance of image enhancement ultimately.
}

\subsection{Data Preparation}

\ly{
To construct our database, we utilize the publicly available LOw-Light database (LOL) \cite{wei2018deep}, which contains 689 low-light and normal-light image pairs for training and 100 low-light and normal-light image pairs for testing. 
According to diverse image content, we selected 247 and 43 low-light images from the LOL training and testing sets, respectively. 
We apply ten classical and state-of-the-art low-light image enhancement methods to enhance the selected low-light images. 
The traditional methods include the Camera Response Model (CRM) \cite{ying2017new}, Joint Enhancement and Denoising Method (JED) \cite{ren2018joint}, Multiscale Retinex (MR) \cite{jobson1997multiscale}, Simultaneous Reflectance and Illumination Estimation (SRIE) \cite{fu2016weighted}, Multiple Fusion (MF) \cite{fu2016fusion}.
The deep learning-based enhancement algorithms tend to introduce different distortions; as such, we add Deep Retinex Decomposition (DRD) \cite{wei2018deep}, ZeroDCE \cite{guo2020zero}, EnlightenGAN \cite{jiang2021enlightengan}, Self-supervised \cite{zhang2020self}, and Deep Recursive Band Network (DRBN) \cite{yang2020fidelity} to enhance the selected low-light images.
The resulting images contain a variety of quality degradations.
As illustrated in Fig.~\ref{fig: Distortion Samples}, these approaches can introduce diverse quality impairments during the enhancement procedure, such as noise contamination, blurring, over-exposure, under-exposure, contrast reduction, color shift, and detail loss. 
Furthermore, it is worth noting that noise can be amplified in the image, resulting in distinct artifacts that are different from synthetic distortions. 
These artifacts can be introduced intentionally or unintentionally and may include mixed distortion patterns that can have a notable influence on the overall quality of the image. 
It is important to consider these types of artifacts when evaluating the performance of different image enhancement methods.

}

\subsection{Subjective Testing}
\begin{figure*} 
\centering    
\subfloat[Input] {\includegraphics[width=0.30\linewidth]{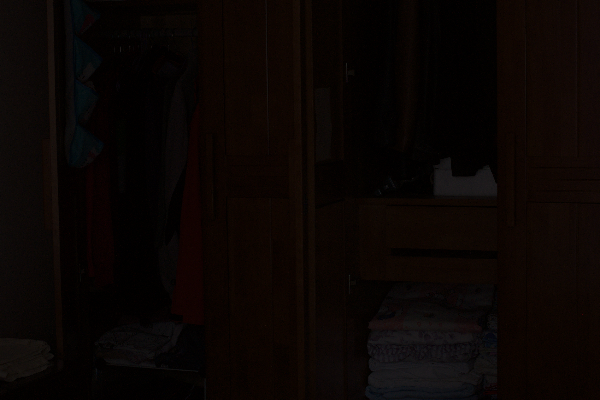}} \hskip0.3em 
\subfloat[JED, OS: 0.1778] {\includegraphics[width=0.30\linewidth]{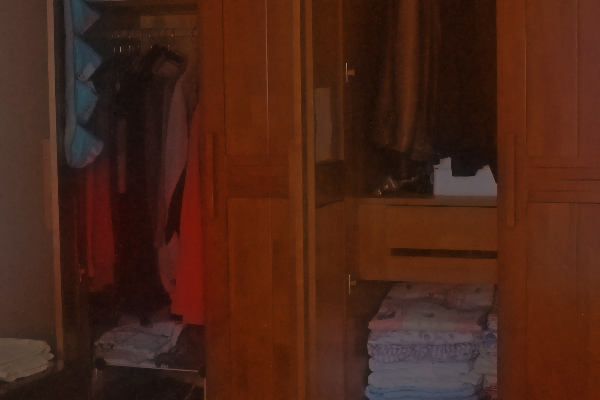}} \hskip0.3em  
\subfloat[Self-supervised ,OS: 0.3667] {\includegraphics[width=0.30\linewidth]{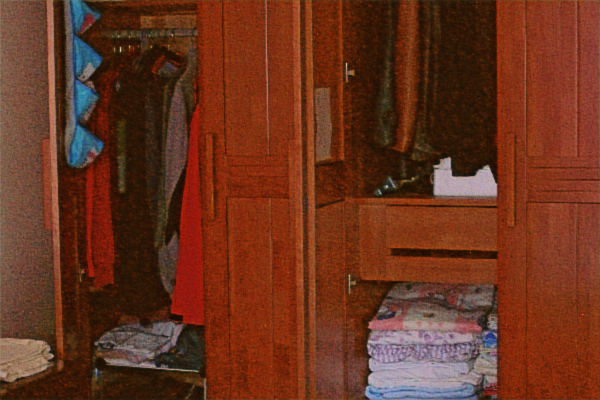}} \hskip0.3em    
\quad
\vskip-1em 
\subfloat[DRD, OS: 0.3704] {\includegraphics[width=0.30\linewidth]{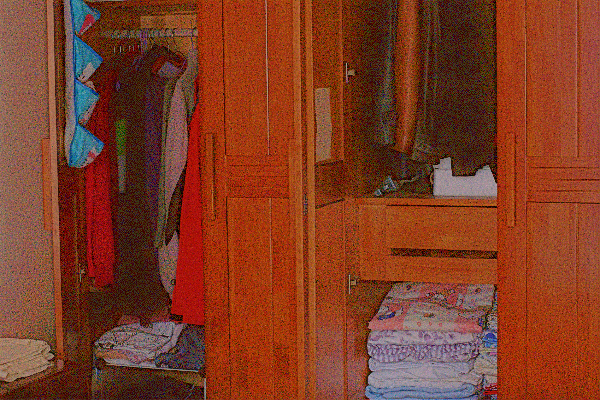}} \hskip0.3em  
\subfloat[DRBN, OS: 0.4519] {\includegraphics[width=0.30\linewidth]{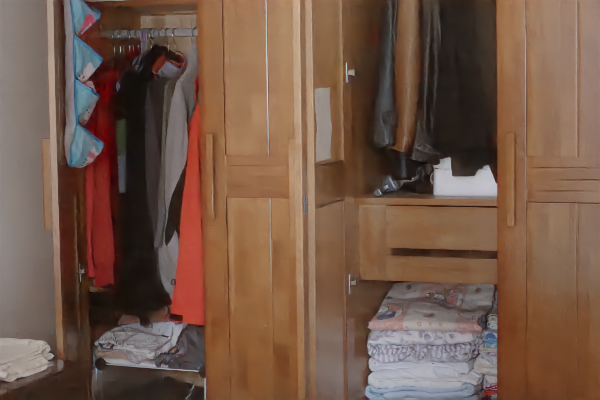}} \hskip0.3em    
\subfloat[SRIE, OS: 0.4630] {\includegraphics[width=0.30\linewidth]{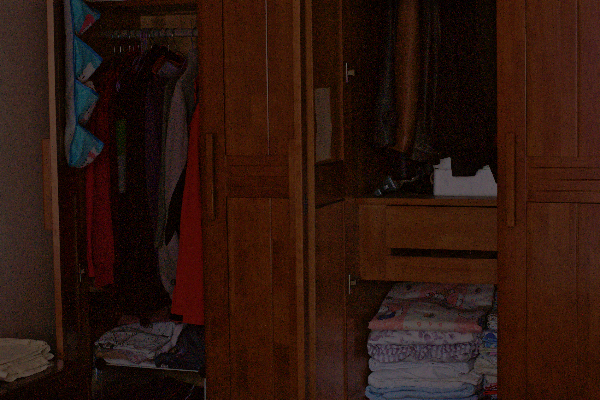}} \hskip0.3em 
\quad
\vskip-1em 
\subfloat[MR, OS: 0.4704] {\includegraphics[width=0.30\linewidth]{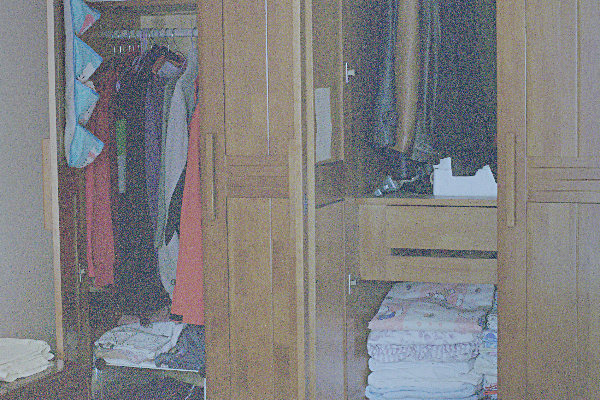}} \hskip0.3em 
\subfloat[MF, OS: 0.5741] {\includegraphics[width=0.30\linewidth]{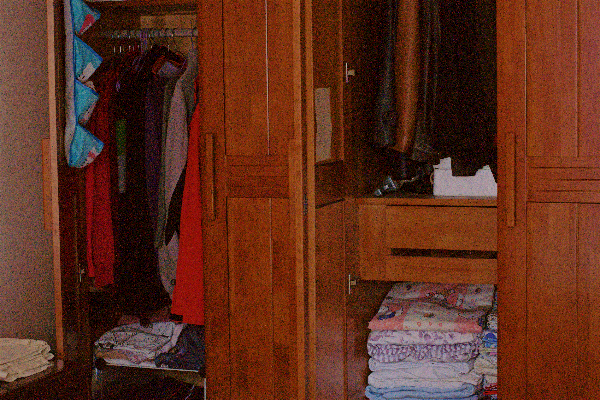}} \hskip0.3em  
\subfloat[Zero, OS: 0.6333] {\includegraphics[width=0.30\linewidth]{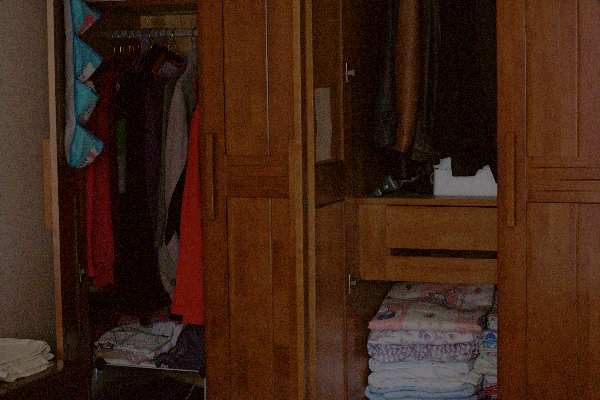}} \hskip0.3em  
\quad
\vskip-1em 
\subfloat[EnlightenGAN, OS: 0.6852] {\includegraphics[width=0.30\linewidth]{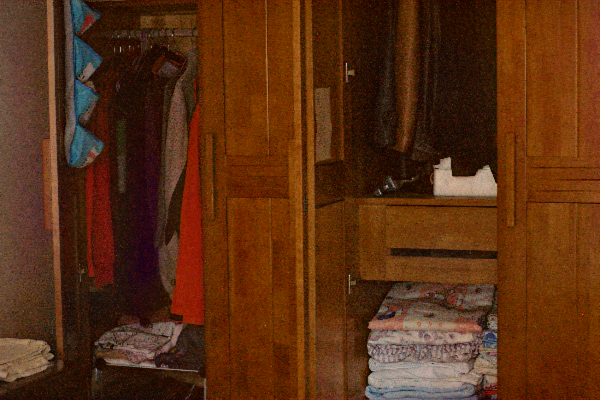}} \hskip0.3em
\subfloat[CRM, OS: 0.8074] {\includegraphics[width=0.30\linewidth]{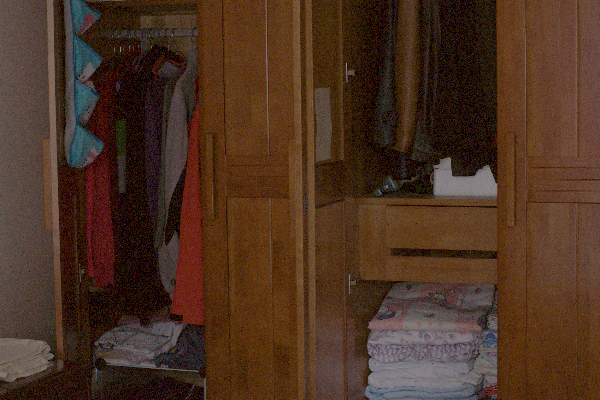}} \hskip0.3em    
\subfloat[Reference ] {\includegraphics[width=0.30\linewidth]{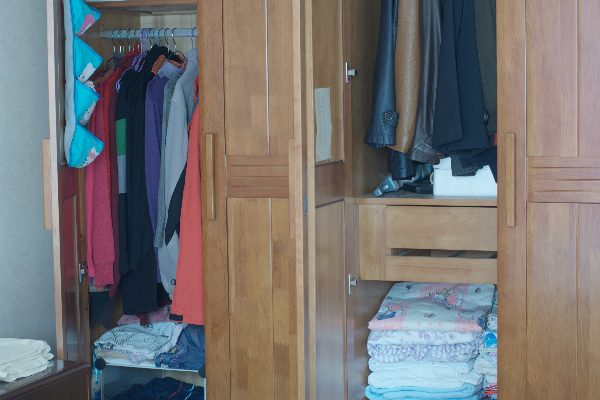}} \hskip0.3em  
\quad
\caption{
\bb{Illustration of the enhancement results with different methods in the SQUARE-LOL database. The corresponding reference image and low-light image are presented in sub-figure (a) and (l), respectively. The opinion score (OS) is provided below each sub-image.}
}
\label{fig:Sample Ranking1}
\end{figure*}

\begin{figure*} 
\centering    
\subfloat[Input] {\includegraphics[width=0.30\linewidth]{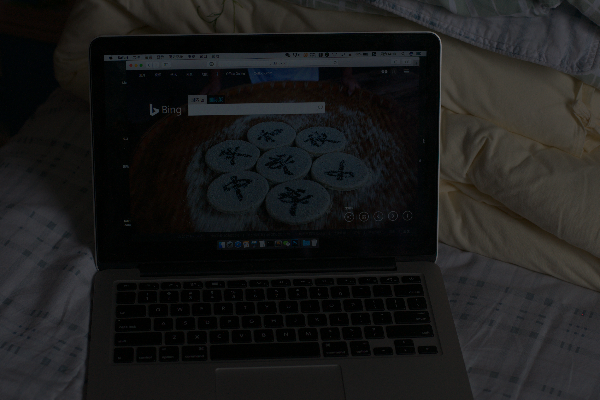}} \hskip0.3em
\subfloat[MR, OS: 0.1704] {\includegraphics[width=0.30\linewidth]{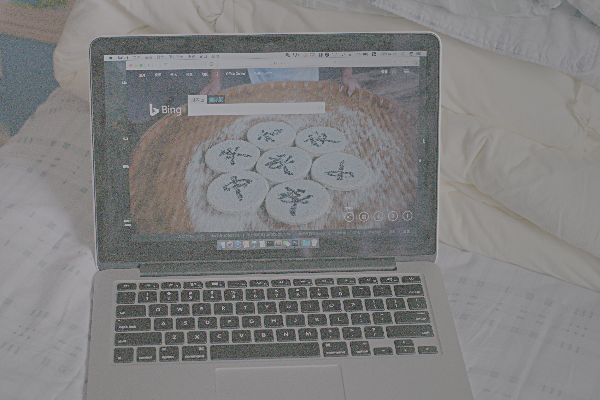}} \hskip0.3em 
\subfloat[DRD, OS: 0.2630] {\includegraphics[width=0.30\linewidth]{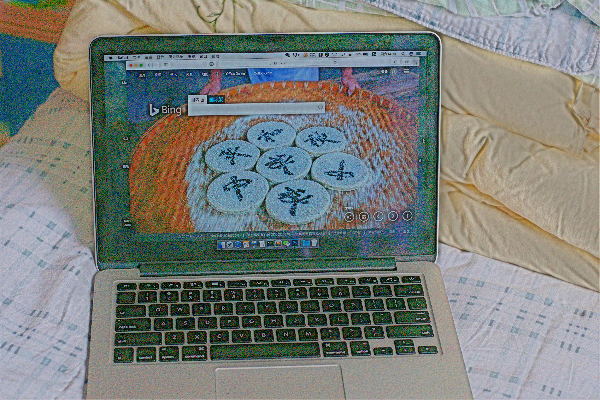}} \hskip0.3em  
\quad
\vskip-1em 
\subfloat[JED, OS: 0.3593] {\includegraphics[width=0.30\linewidth]{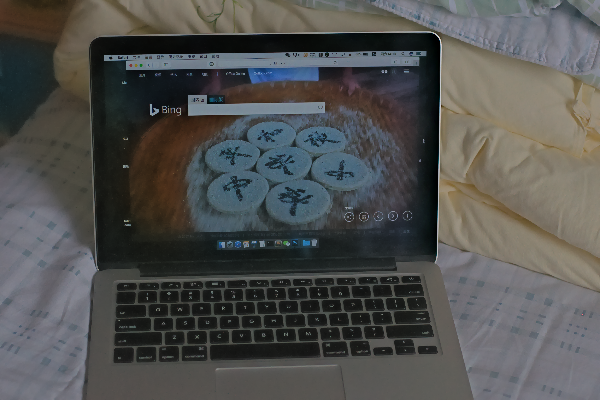}} \hskip0.3em  
\subfloat[Self-supervised, OS: 0.3667] {\includegraphics[width=0.30\linewidth]{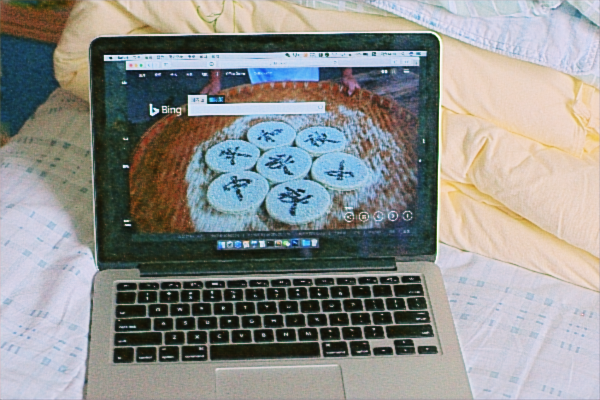}} \hskip0.3em  
\subfloat[DRBN, OS: 0.4926] {\includegraphics[width=0.30\linewidth]{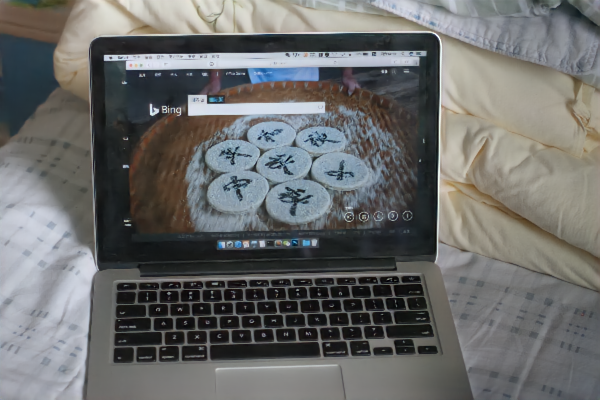}} \hskip0.3em    
\quad
\vskip-1em 
\subfloat[SRIE, OS: 0.5444] {\includegraphics[width=0.30\linewidth]{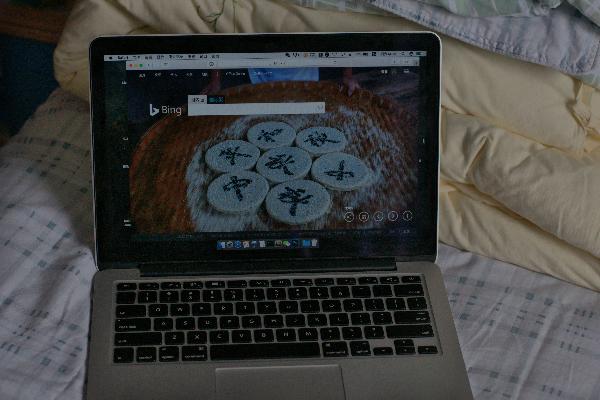}} \hskip0.3em 
\subfloat[CRM, OS: 0.6074] {\includegraphics[width=0.30\linewidth]{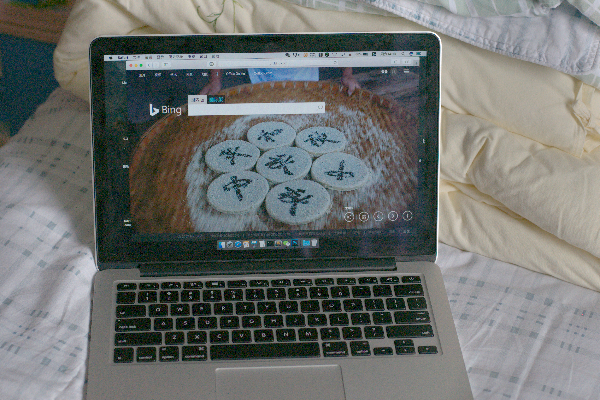}} \hskip0.3em  
\subfloat[Zero, OS: 0.6148] {\includegraphics[width=0.30\linewidth]{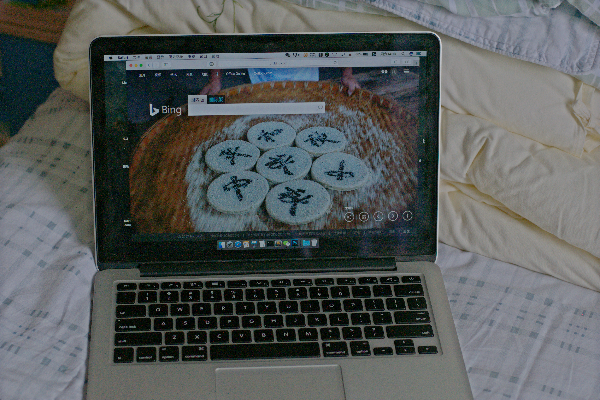}} \hskip0.3em  
\quad
\vskip-1em 
\subfloat[EnlightenGAN, OS: 0.7185] {\includegraphics[width=0.30\linewidth]{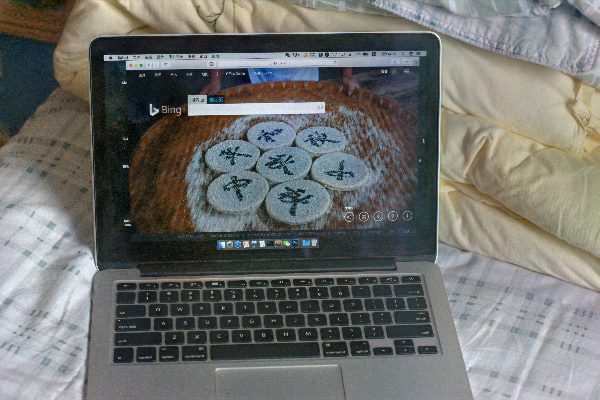}} \hskip0.3em  
\subfloat[MF, OS: 0.8000] {\includegraphics[width=0.30\linewidth]{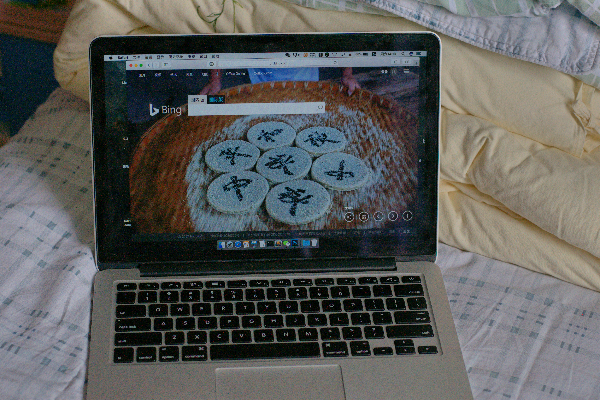}} \hskip0.3em  
\subfloat[Reference ] {\includegraphics[width=0.30\linewidth]{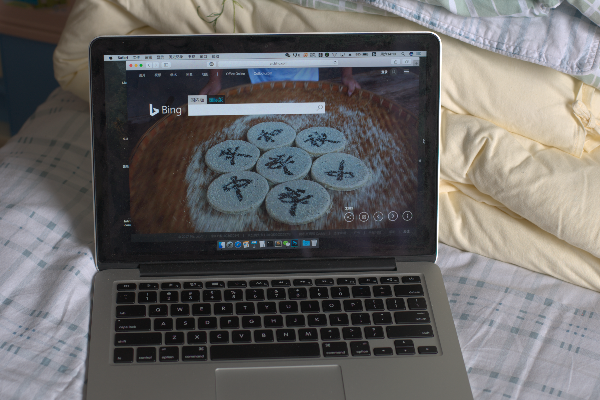}} \hskip0.3em  
\quad
\caption{
\bb{Illustration of the enhancement results with different methods in the SQUARE-LOL database. The corresponding reference image and low-light image are presented in sub-figure (a) and (l), respectively. The opinion score (OS) is provided below each sub-image.}}
\label{fig:Sample Ranking2}
\end{figure*}

\ly{
The primary objective of this work is to conduct a detailed analysis of the perceptual visual quality of enhanced results and the behavior of low-light image enhancement algorithms. 
To achieve this goal, we have created a dedicated database that focuses exclusively on pairwise preference. 
In other words, each trial in the database presents two enhanced images to the subject, who must choose the one with a more pleasing visual quality.
This method is commonly known as the \textbf{Two}-\textbf{A}lternative \textbf{F}orced \textbf{C}hoice (\textbf{2AFC}) and is widely used in psychophysical studies due to its high discrimination power, which leads to more trustworthy subjective quality score.
}
In detail, each subject attends a tutorial with detailed instructions before the subjective experiment. The subjects are required to identify the image with better quality within 2$\sim$5 seconds. 
We use LCD monitors (resolution: $ 1920 \times 1080$ pixels) to perform subjective experiments. These experiments are calibrated according to recommendations in~\cite{bt2002methodology}. 
The subjective experiments are conducted with random display pairs, and each pair is generated with an identical low-light image but different enhancement models. As such, the total number of pairs is $C_{10}^{2} \times 290=13,050.$
%
We follow \cite{lai2016comparative} and adopt the sanity check to ensure the accuracy of the comparison after each session. 
In total, \(N\) subjects are invited in the subjective testing, and herein \(N\)=30.  
Eventually,  for each enhanced image, we are able to collect $9 \times 30 =270$ votes from the 30 subjects. 
In Fig.~\ref{fig:Sample Ranking1} and Fig.~\ref{fig:Sample Ranking2}, we show two scenes sampled from our SQUARE-LOL database, and the corresponding subjective testing results are shown in Table~\ref{tab: example mos}. 
From  Table~\ref{tab: example mos}, we can observe that the image enhanced by EnlightenGAN is assigned 197 votes and acquire the highest opinion score ($197/270 = 0.7296$) among the ten enhancement algorithms, revealing the superior perceptual quality. 

\begin{table*}[]
\caption{ Pairwise comparison results provided by 30 subjects. 
The reported values correspond to the number of times that the method in the row was preferred over the method in the column, indicating its relative superiority. 
WT denotes Winning Times, and OS denotes Opinion Score.}
\label{tab: example mos}
\centering
\begin{tabular}{c|cccccccccc|ccc}
\hline\hline
                & CRM   & EnlightenGAN  & JED  & MF  & MR   & DRD   & Self-supervised    & DRBN   & SRIE  & ZeroDCE  & WT  & Total  & OS\\\hline
CRM             &-      & 14  & 19   & 18   & 29  & 30    & 19   & 15     & 22    & 19      & 185  & 270   & 0.6852 \\
EnlightenGAN    &16     & -   & 22   & 22   & 30  & 30    & 29   & 15     & 19    & 14      & 197  & 270   & 0.7296 \\
JED             & 11    & 8   & -    & 9    & 18  & 23    & 15   & 9      & 14    & 11      & 118  & 270   & 0.4370 \\
MF              &12     & 8   & 21   & -    & 30  & 30    & 22   & 14     & 16    & 13      & 166  & 270   & 0.6148 \\
MR              & 1     & 0   & 12   & 0    & -   & 30    & 11   & 11     & 5     & 0       & 70   & 270   & 0.2593 \\
DRD             & 0     & 0   & 7    & 0    & 0   & -     & 2    & 9      & 2     & 2       & 22   & 270   & 0.0815 \\
Self-supervised & 11    & 1   & 15   & 8    & 19  & 28    & -    & 15     & 7     & 3       & 107  & 270   & 0.3963\\
DRBN            & 15    & 15  & 21   & 16   & 19  & 21    & 15   & -      & 2     & 11      & 135  & 270   & 0.5000 \\
SRIE            & 8     & 11  & 16   & 14   & 25  & 28    & 23   & 28     & -     & 14      & 167  & 270   & 0.6185 \\
ZeroDCE         & 11    & 16  & 19   & 17   & 30  & 28    & 27   & 19     & 16    & -       & 183  & 270   & 0.6778 \\
\hline\hline
\end{tabular}
\end{table*}

\subsection{Data Analyses}
\begin{figure}[t]
\begin{minipage}[b]{1.0\linewidth}
  \centering
  \centerline{\includegraphics[width=1\linewidth]{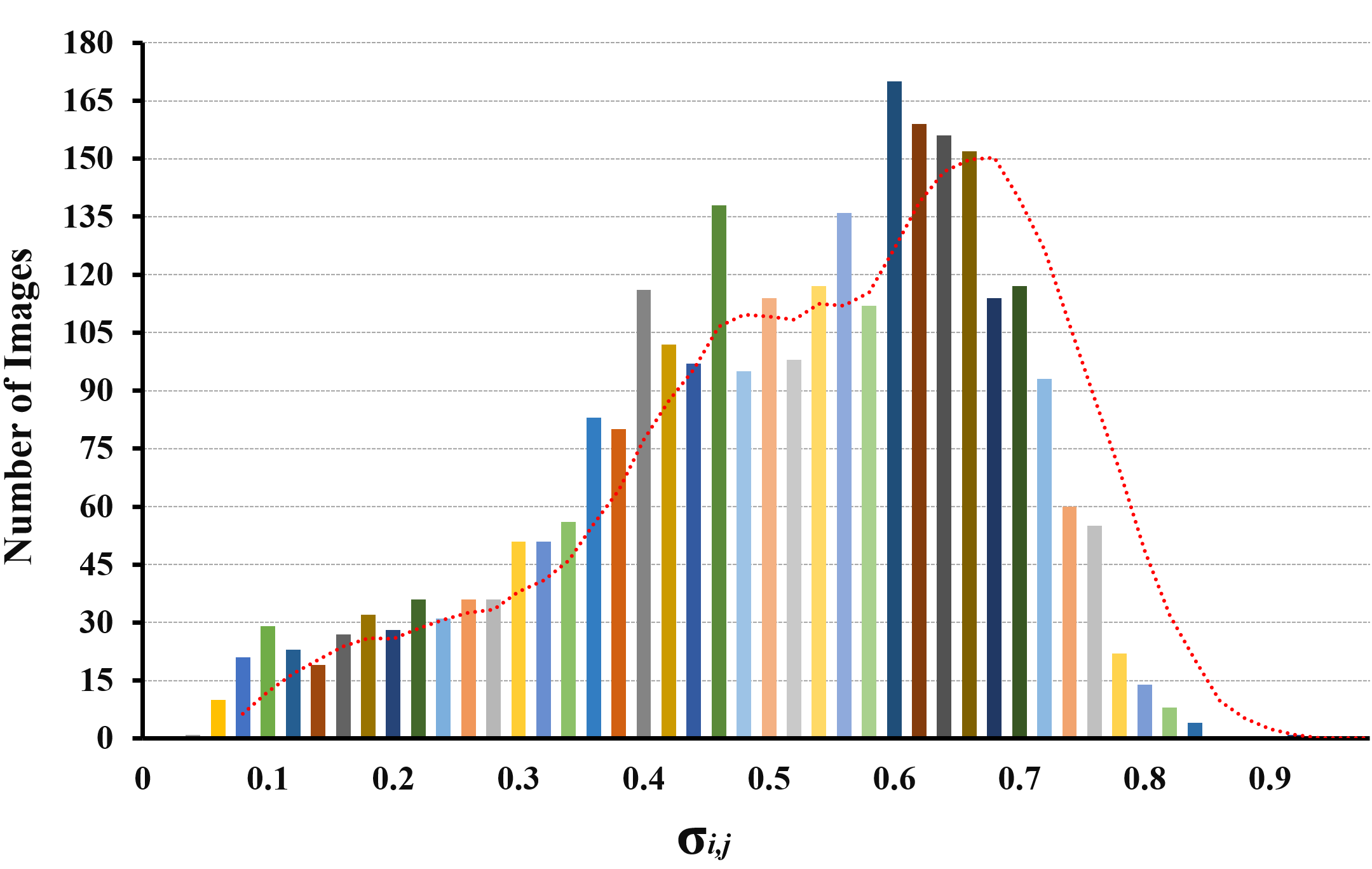}}
\end{minipage}
\caption{\bb{The distribution of opinion scores of our SQUARE-LOL database.}
}
\label{fig:mos-distribution}
\end{figure}

To quantify the image quality, we assign ``1'' to the particular image when it is preferred in the pairwise comparison for each trail.  
Given one image $I_{i,j}$, where \(i\) represents the image index and \(j\) denotes the enhancement model, in total, there are \(T=N\times (M-1)\) pairwise comparisons where \(N\) and \(M\) are the numbers of subjects and models, respectively.
Assuming that the image is in favor of for \(V_{i,j}\) times, the ground-truth opinion score that quantifies the quality with the particular enhancement model is given by,
\begin{equation}
\sigma_{i,j}=\frac{V_{i,j}}{T}.
\end{equation}
\ly{The distribution of \(\sigma_{i,j}\) is presented in Fig.~\ref {fig:mos-distribution}, indicating that our database covers a broad spectrum of quality levels. 
However, it is worth noting that there are very few enhanced images within the range of [0.8, 1.0]. 
This can be attributed to the lack of a universal algorithm with high generalization capability for images with diverse content. 
The proposed database, with its subjective ratings, offers valuable insights into the development of more effective enhancement systems.}

\begin{figure*}[ht]
\begin{center}
\includegraphics[width=1.0\textwidth]{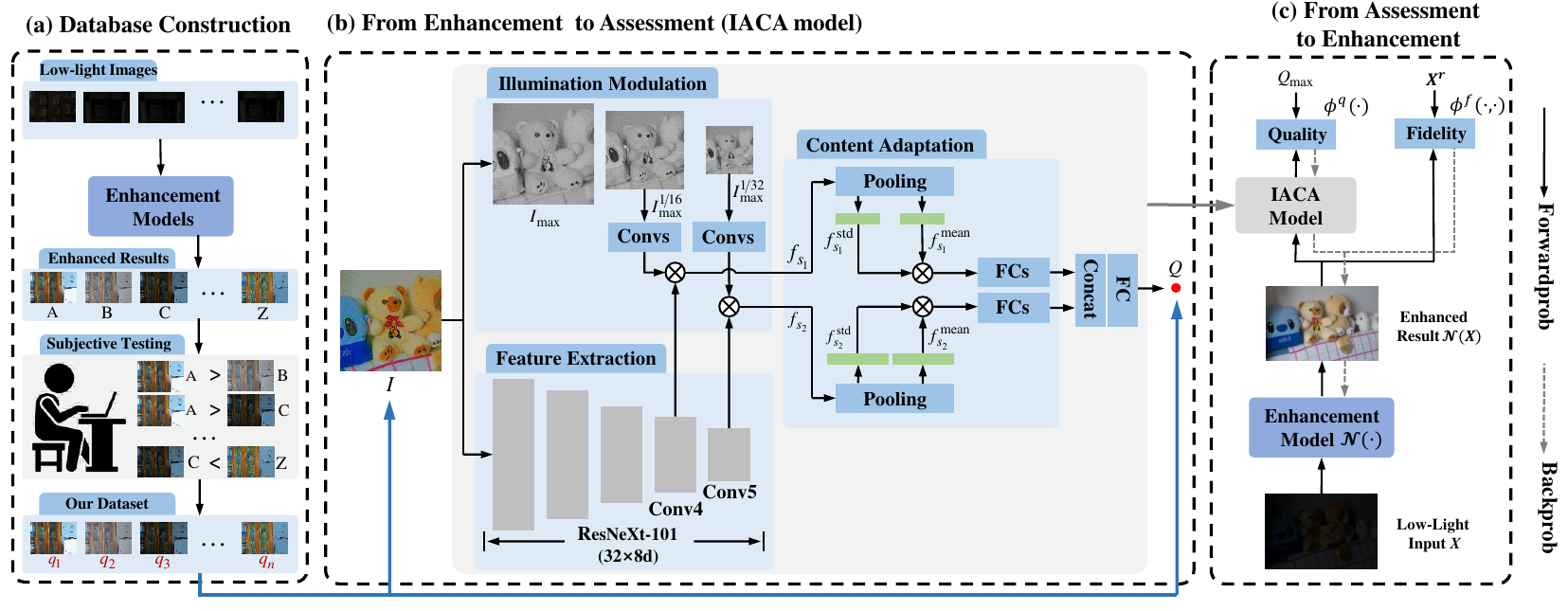}
\caption{
\ly{Illustration of the proposed comprehensive scheme to bridge the gap between quality assessment and enhancement for low-light images.
This scheme can be summarized as follows:
a) The construction of the valuable \textit{SQUARE-LOL database}, which provides enhanced low-light images generated using different enhancement (traditional and deep-learning) methods with corresponding quality labels; 
b) The development of a deep learning-based IQA measure (IACA) dedicated to enhanced images by bridging the gap \textit{from enhancement to assessment}; 
c) The optimization of enhancement models with the guidance of the IACA model, filling in the gap \textit{from assessment to enhancement}. 
}
}
\label{fig:arc}
\end{center}
\end{figure*}

\section{Methodology}
%
\ly{
As depicted in Fig.~\ref{fig:arc}, this operation chain is bridged by the proposed IQA model, which plays an essential role in ensuring that the final enhanced images meet the essential requirement of perceptual quality.
The approach involves learning to enhance low-light images using a full operation chain that transfers knowledge from subjective quality assessment to the enhancement process. 
To accomplish this, we develop the IACA (Image Assessment with Convolutional Attention) model, which is trained using the SQUARE-LOL database. Using this IQA model, we then aim to learn how to enhance low-light images, ultimately creating visually more pleasing results.
}

\subsection{From Enhancement to Assessment}
By leveraging the established SQUARE-LOL database, we propose a deep learning-based IQA model (IACA)  for low-light enhancement images. 
We show the architecture of our deep network in Fig.~\ref{fig:arc}. In our model, we adopt ResNeXt
\cite{xie2017aggregated} as the backbone for feature extraction. 
Given an image \textit{$I$}, it is passed through the CNN (ResNeXt - $32 \times 8d$  \cite{xie2017aggregated}) for multi-scale feature extraction. 
In particular, the  ResNeXt - $32 \times 8d$ is pre-trained on ImageNet \cite{deng2009imagenet}, and the final fully-connected layer is discarded.  
%
%
However, the illumination information, which plays a critical role in the low-light enhancement, has not been specifically considered in the pre-trained ResNeXt. 
As such, the illumination modulation strategy is incorporated into our model. Moreover, to further enhance the generalization capability, a content adaptation module based on channel attention is employed for feature aggregation.
%

\subsubsection{Illumination  Modulation} 
There are several reasons that account for the ignorance of illumination when the vanilla ResNeXt model is adopted for feature extraction. 
First, the ResNeXt is pre-trained for object classification, expecting to be robust and invariant even under different illumination conditions. 
Second, the data used for training in ImageNet are usually images with normal light. 
In view of this, we first estimate the luminance map of input images with the Max-RGB strategy \cite{hussain2016max}, such  that the highest pixel values in each  channel (R, G, B) of the input image can be extracted, which are further denoted as \textit{$I_{max}$}.
For each scale, we adopt max pooling layer to pool \textit{$I_{max}$} into the luminance map with a spatial size identical to the corresponding feature map. 
Following the pooled luminance map, three convolutional layers are adopted to extract the illumination-aware features, which are merged with the feature maps from the original images. 
Following this vein, the luminance information can be actively involved in the model learning process. 

\subsubsection{Content Adaptation} 
%

\ly{
To encode the generalization capability into feature representation, we leverage statistical pooling moments and aggregate the features in the fourth layer $Conv4$ and the fifth layer $Conv5$ with global average pooling (mean pooling) and standard deviation pooling (std pooling) layers. 
In Fig.~\ref{fig:arc}, we use the notation $f_{s1}^{mean}$, $f_{s1}^{std}$, $f_{s2}^{mean}$, and $f_{s2}^{std}$ to represent the aggregated features from layer $Conv4$ and $Conv5$. 
Specifically, $f_{s1}^{mean}$ and $f_{s1}^{std}$ denote the mean and standard deviation of the features from $Conv4$, while $f_{s2}^{mean}$ and $f_{s2}^{std}$ denote the mean and standard deviation of the features from $Conv5$. 
Simply concatenating the pooled features $f_{s1}^{mean}$ and $f_{s2}^{mean}$ for quality regression may not be practical. 
This is because these features are highly correlated with semantic information \cite{wan2019information,chen2021learning}, which can cause overfitting to specific scenes in the training set. 
Instead of discarding $f_{s1}^{mean}$ and $f_{s2}^{mean}$, we treat them as semantically meaningful features working as an integral part of the attention-based multi-scale feature extraction.
Specifically, we adopt four Fully-Connected (FC) layers with the following units: 256, 64, 256, 1024, and 256, 64, 256, 2048 for $f_{s1}^{mean}$ and $f_{s2}^{mean}$, respectively. For the first three FC layers, each FC is followed by a Rectified Linear Unit (ReLU) activation layer \cite{glorot2011deep}.
}
For the last layer, the activation layer is Sigmoid, by which the attention map can be estimated.  
Subsequently, the attention weighted \textit{$f_{s1}^{std}$} and \textit{$f_{s2}^{std}$} are further encoded by three FC layers, then we concatenate the encoded features and use the FC layer for quality regression.

\begin{table*}
  \centering
  \caption{Comparisons of quality prediction performance on SQUARE-LOL. The values corresponding to the best and second-best performance are highlighted in boldface and underlined, respectively.}

\begin{tabular}{c|cccc|cccc|c}
\hline
\multirow{2}{*}{Models} & \multicolumn{4}{c|}{Conventional Methods} & \multicolumn{4}{c|}{Deep-learning Based Methods} & \multirow{2}{*}{\textbf{IACA (Ours) }} \\
\cline{2-9}      & NIQE  & BRISQUE & CORNIA & HOSA & Bilinear & WaDIQaM-NR & Normal-In-Normal & Meta-IQA &  \\
\hline
SROCC & 0.276       &0.658       & 0.625      & 0.685      & 0.792  & 0.838  &\underline{0.850}      & 0.654    & \textbf{0.875}  \\
PLCC  & 0.286      &0.649       & 0.631      & 0.694      & 0.783  & 0.841  & \underline{0.844}     & 0.668    & \textbf{0.878}  \\
\hline
\end{tabular}%

 \label{tab:iqa}%
\end{table*}%


\subsection{From Assessment to Enhancement}




\ly{
The underlying philosophy of many low-light image enhancement technologies is to recover images using the Maximum A Posteriori (MAP) framework. 
This approach involves simultaneously maintaining fidelity (fidelity term) and the alignment of scene statistics (prior term)~\cite{banham1997digital}.
By incorporating both fidelity and prior terms, the MAP framework can effectively enhance low-light images while preserving important scene information.
}
In our optimization framework, we follow this vein and adopt the widely used full-reference IQA model structural similarity (SSIM) index \cite{wang2004image} for signal-level comparisons in an effort to push the image to be closer to the ground truth in supervised learning. 
More importantly, regarding the prior term, instead of adopting hand-crafted image priors such as dark channel~\cite{he2010single}, 
Total Variation (TV)~\cite{chen2018variational} or Markov random field (MRF)~\cite{li2009markov}, the valid IACA model in the perceptual sense is incorporated.  
Thus, we can define the loss function for learning the model in the following way,
\begin{equation}\label{losscom}
\mathcal{L} = \phi^{f}\left(\mathbf{X}^{r}, \mathcal{N}(\mathbf{X}; \mathbf{W})\right)+\lambda \phi^{q}\left(\mathcal{N}(\mathbf{X}; \mathbf{W})\right),
\end{equation}
where the enhancement network and its parameters are denoted as $\mathcal{N}$ and $\mathbf{W}$, respectively. 
And the notation $\mathbf{X}$ and $\mathbf{X}^{r}$ is used to refer to the input low-light image and its corresponding reference image, respectively
The fidelity loss term $\phi^{f}(\cdot,\cdot)$ is based on SSIM, and the quality loss term  $\phi^{q}(\cdot)$ is given by,
\begin{equation}\label{losscom2}
\phi^{q}(\mathcal{N}(\mathbf{X}; \mathbf{W})) = \left|Q^{max} - IACA(\mathcal{N}(\mathbf{X}; \mathbf{W}))\right|,
\end{equation}
where $Q^{max}$ denotes the maximum quality score the IACA model can achieve on the training set.  
The hyper-parameter $\lambda$ in Eqn.~\eqref{losscom} is used to adjust the relative importance of the IACA model in the loss function.

To accurately represent the intrinsic performance of model optimization, we employ the SRCNN \cite{dong2015image} framework and the DRBN \cite{yang2020fidelity} architecture, which is considered state-of-the-art in low-light enhancement, for model learning. Specifically, we substitute the original loss functions utilized in these two models with the proposed quality-aware loss function (see Eqn.~\eqref{losscom}).
%
For fair comparisons, we fix the network architecture of the two models, including kernel size, layer numbers, and weight initialization. Moreover, the baseline is generated by the removal of the term guided by IACA in the loss function. 
%


\begin{figure*}[!ht] 
  \centering 
  \begin{minipage}[b]{1.0\linewidth} 
  \subfloat[SRCNN Ori.]{
    \begin{minipage}[b]{0.21\linewidth} 
      \centering
      \includegraphics[width=\linewidth]{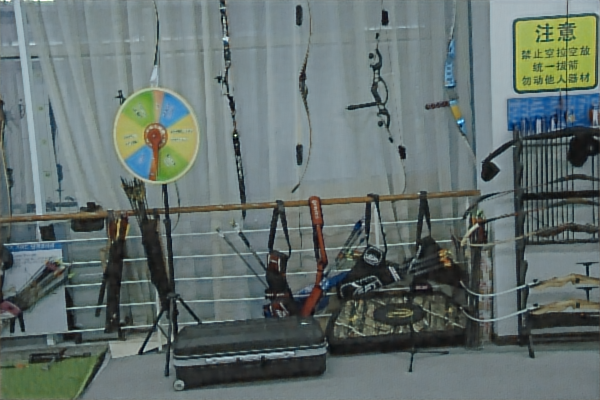}\vspace{1pt}
      \includegraphics[width=\linewidth]{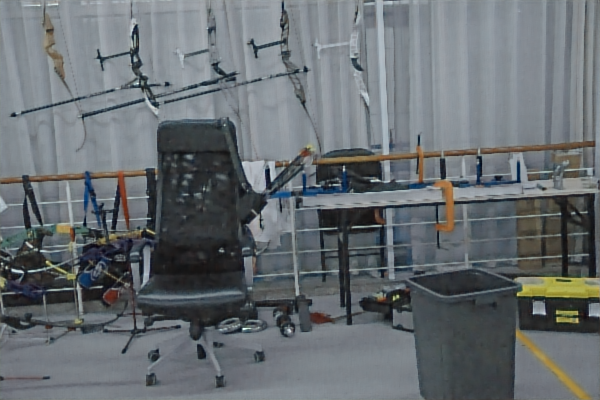}\vspace{1pt}
      \includegraphics[width=\linewidth]{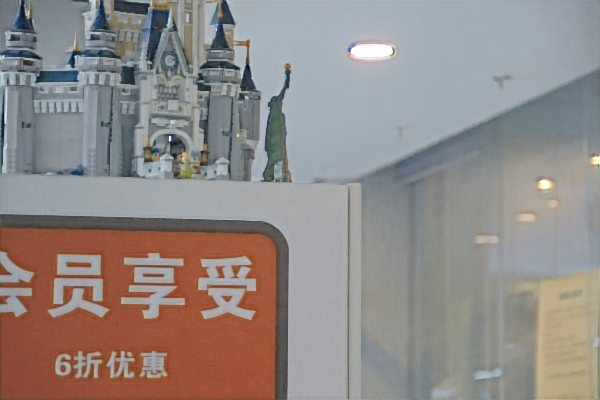}\vspace{1pt}
      \includegraphics[width=\linewidth]{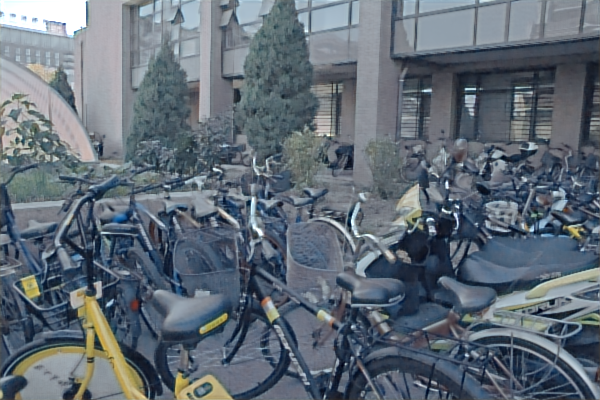}\vspace{1pt}
      \includegraphics[width=\linewidth]{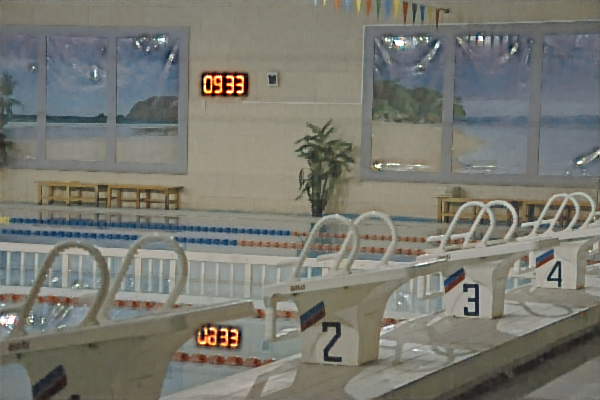}
    \end{minipage}
  }
  \hspace{.05in}
   \subfloat[SRCNN Opt.]{
    \begin{minipage}[b]{0.21\linewidth}
      \includegraphics[width=\linewidth]{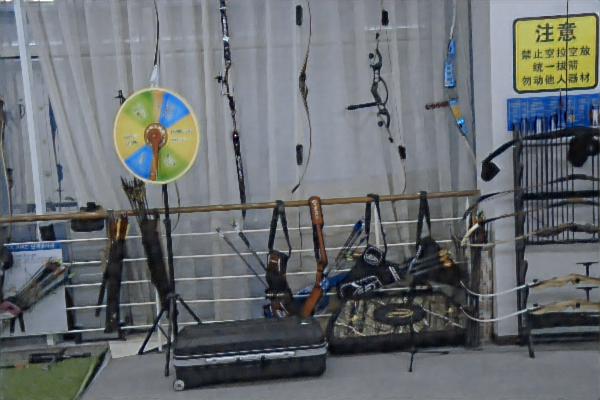}\vspace{1pt}
      \includegraphics[width=\linewidth]{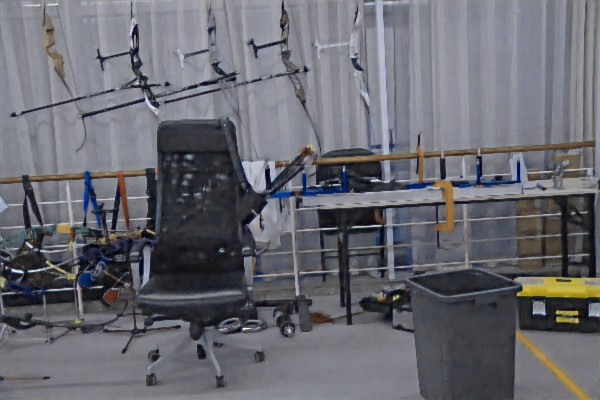}\vspace{1pt}
      \includegraphics[width=\linewidth]{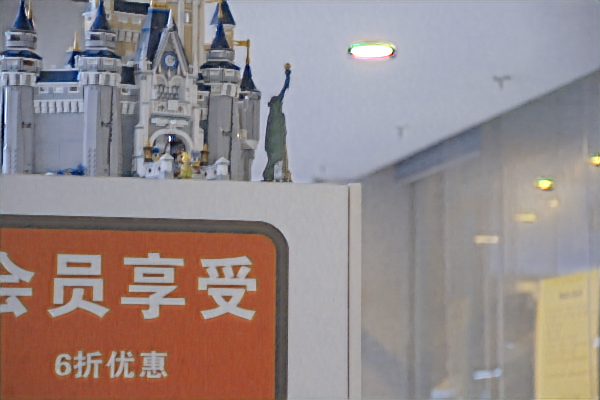}\vspace{1pt}
      \includegraphics[width=\linewidth]{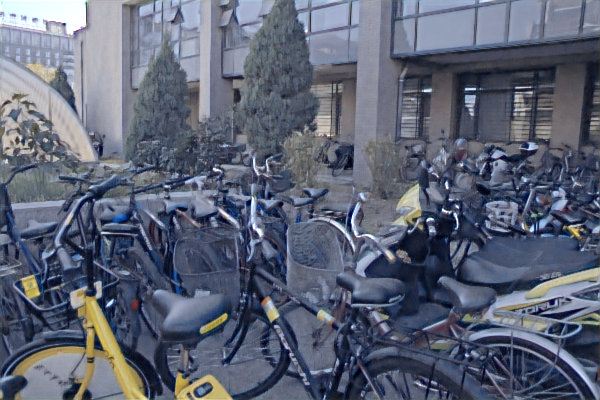}\vspace{1pt}
      \includegraphics[width=\linewidth]{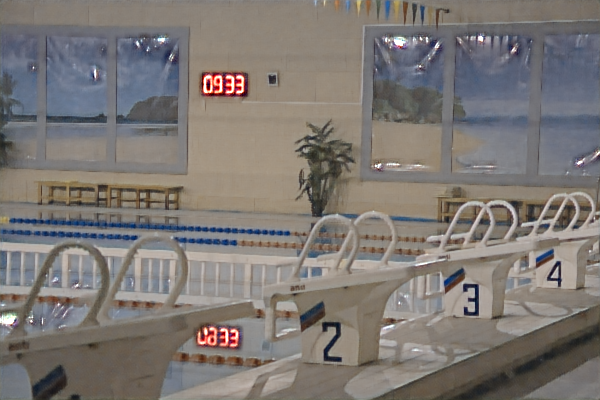}
    \end{minipage}
  }
  \hspace{.05in}
    \subfloat[DRBN Ori.]{
    \begin{minipage}[b]{0.21\linewidth}
      \centering
      \includegraphics[width=\linewidth]{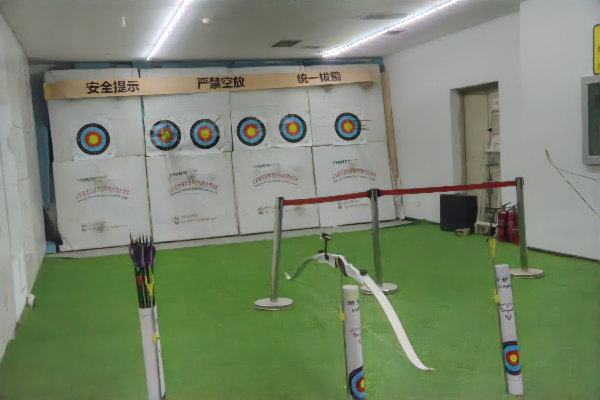}\vspace{1pt}
      \includegraphics[width=\linewidth]{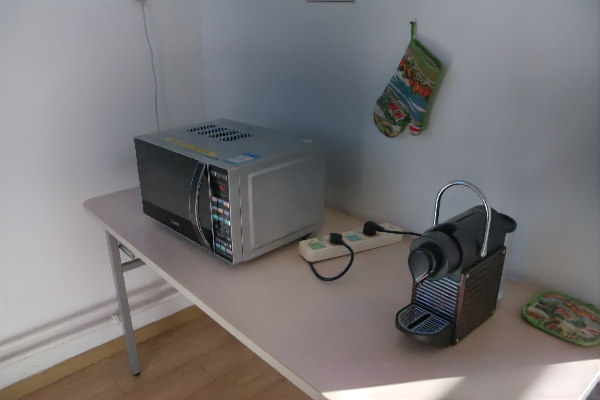}\vspace{1pt}
      \includegraphics[width=\linewidth]{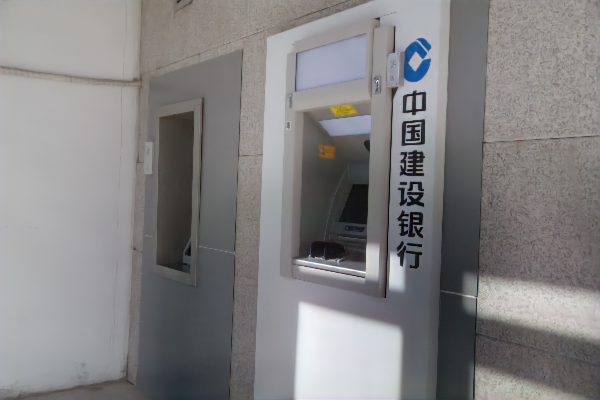}\vspace{1pt}
      \includegraphics[width=\linewidth]{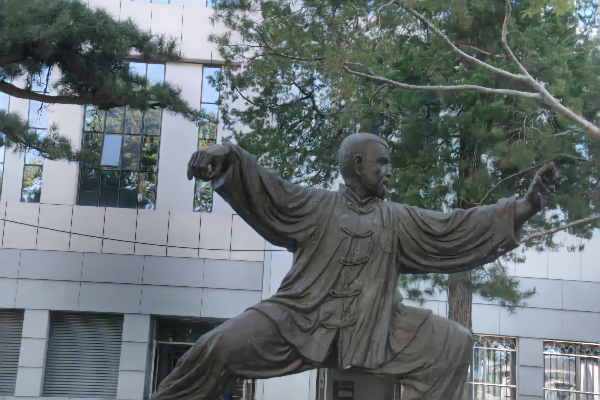}\vspace{1pt}
      \includegraphics[width=\linewidth]{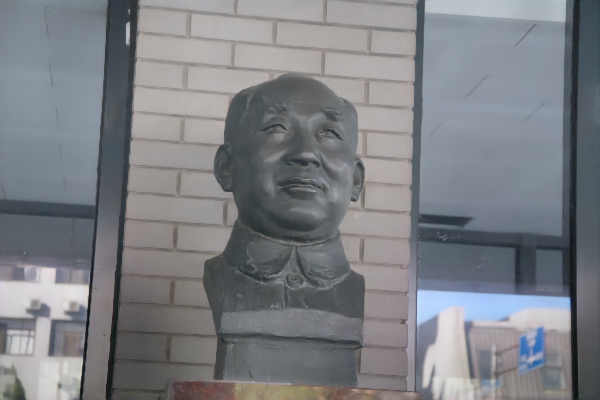}
    \end{minipage}
  }
 \hspace{.05in}
    \subfloat[DRBN Opt.]{
    \begin{minipage}[b]{0.21\linewidth}
      \centering
      \includegraphics[width=\linewidth]{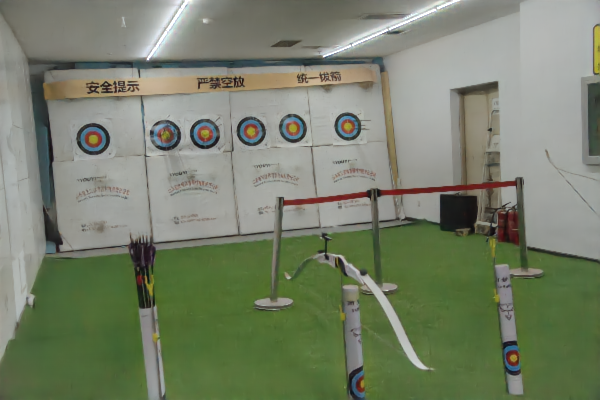}\vspace{1pt}
      \includegraphics[width=\linewidth]{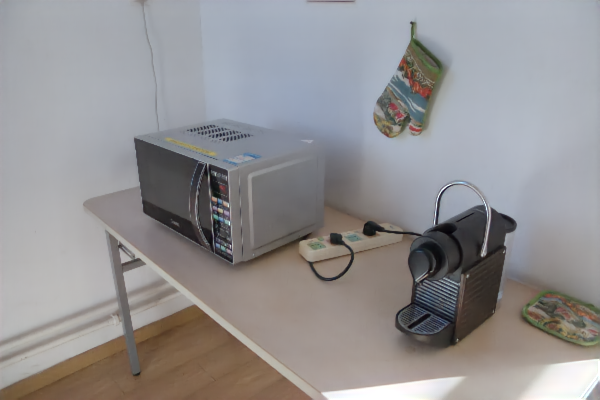}\vspace{1pt}
      \includegraphics[width=\linewidth]{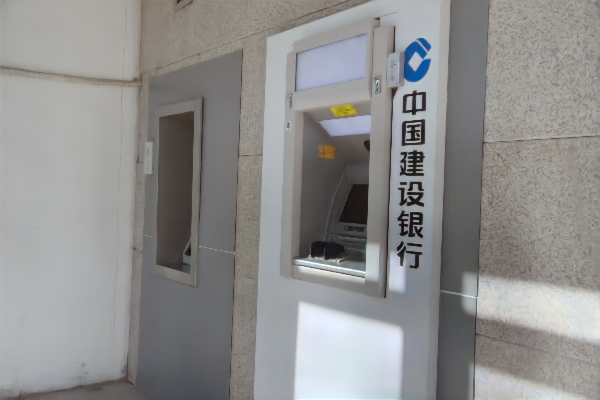}\vspace{1pt}
      \includegraphics[width=\linewidth]{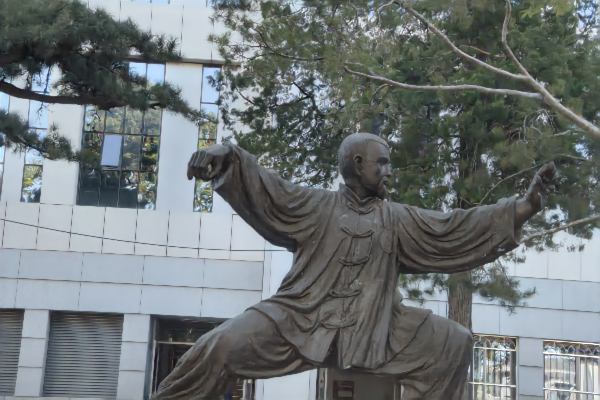}\vspace{1pt}
      \includegraphics[width=\linewidth]{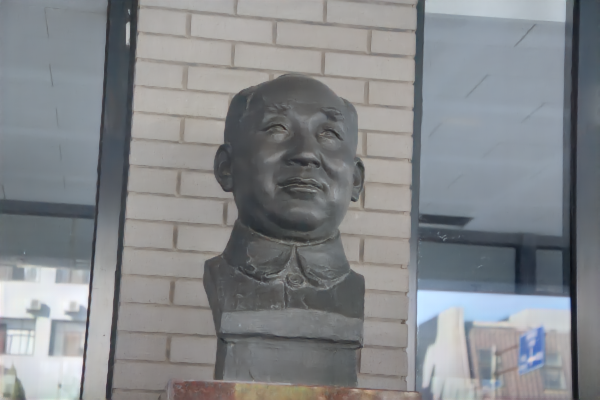}
    \end{minipage}
  }
  \end{minipage}
  \vfill
  \caption{Visual quality comparisons of enhanced images by original baselines (SRCNN Ori. and DRBN Ori.) and our optimized models (SRCNN Opt. and DRBN Opt.). }
  \label{fig:cmpr}
\end{figure*}
\section{Experiments}

\subsection{Experimental Settings}
During the IACA model learning, we divide our database into a training set (2,470 images) and a testing set (430 images) according to the content, and this is consistent with the LOL database~\cite{wei2018deep} to ensure no content overlapping. 
%
%
To increase the diversity of the training data, we randomly crop patches with dimensions of $224 \times 224$ from the original images. 
During optimization, we use a batch size of 32 and apply the Adam optimizer with a learning rate of 1e-4. 
Training is terminated after 100 epochs.
In the testing phase, the images are tested at the original size without cropping. 
For the faster convergence of our model, the ``Norm-In-Norm''~\cite{li2020norm} loss is adopted for regression. 
The loss function is performed not only on the final concatenated features but also on the features in the two sub-scales, and their weights are set to be identical. 
We use the Spearman Rank Order Correlation Coefficient (SROCC) and the Pearson Linear Correlation Coefficient (PLCC) to evaluate the IACA model.


To train the enhancement model, we adopt the default data splitting strategy from the LOL database~\cite{wei2018deep}, which involves creating distinct training and testing sets. 
Notably, this approach ensures that the testing sets and training sets used in this paper do not overlap, either in terms of IACA prediction or enhancement optimization.

As part of the training process, we perform data augmentation by randomly cropping the images to dimensions of $256\times 256$, while the original sizes are used during testing. A total of 200 epochs are allowed during training, with the initial learning rate set to 1e-4. After the 200 epochs, the learning rate is decreased by 0.5. Furthermore, we set the hyper-parameter $\lambda$ in Eqn.~\eqref{losscom} to be 5e-3 in SRCNN and 1e-3 in DRBN, respectively.

\subsection{Experimental Results}
\subsubsection{Quality Prediction Performance Evaluation}


This subsection presents a comprehensive comparison of our proposed IACA model with state-of-the-art BIQA methods, including hand-crafted feature-based (NIQE~\cite{mittal2012making}, BRISQUE~\cite{ mittal2012no}, CORNIA~\cite{ye2012unsupervised}, HOSA~\cite{xu2016blind}) and deep learning-based (Bilinear~\cite{zhang2018blind}, WaDIQaM-NR~\cite{bosse2017deep}, Meta-IQA~\cite{zhu2020metaiqa} and Normal-In-Normal~\cite{li2020norm}) models. 
For the hand-crafted feature-based methods, we directly 
evaluate the models on our SQUARE-LOL database with open-source codes. 
Regarding the deep learning-based methods, we follow the parameter initialization of those models and retrain them on our database using the same training set as the proposed IACA model.


Table~\ref{tab:iqa} clearly demonstrates that the performance of conventional BIQA  methods is significantly below the desired level. 
Due to the intrinsic differences between the enhanced low-light images and the normal-light images, the readily deployed IQA models may not suffice in the low-light enhancement scenarios. 
It is also worth mentioning that ``Norm-In-Norm'' shares the same backbone with ours with only the mean pooling utilized for feature aggregation, causing the model to be less effective. 
Compared with the above deep learning-based methods, we introduce the illumination  modulation in the feature extraction 
as well as the attention strategy, leading to a more generalized model. 
The superior performance validates the effectiveness of the proposed IACA  method. 


\begin{table*}
  \centering
  \caption{Ablation study of each component in the IACA model.}
\begin{tabular}{ccccc|cc}
\hline
MS-Feat & Illumination Modulation & Content Adaptation & Std Pooling & Mean Pooling & SROCC & PLCC \\
\hline
      &  \checkmark     &  \checkmark     &       &       & 0.864 & 0.865 \\
 \checkmark     &       &  \checkmark    &       &       & 0.866 & 0.871 \\
 \checkmark     &  \checkmark     &       &  \checkmark    &       & 0.859 & 0.866 \\
 \checkmark     &  \checkmark    &       &       &  \checkmark     & 0.861 & 0.866 \\
 \hline
 \checkmark     &  \checkmark     &  \checkmark    &       &        & \textbf{0.875}       & \textbf{0.878} \\
\hline
\end{tabular}%

 \label{tab:abl}%
\end{table*}%

\begin{figure*}[htbp]
\begin{center}
\includegraphics[width=0.9\textwidth]{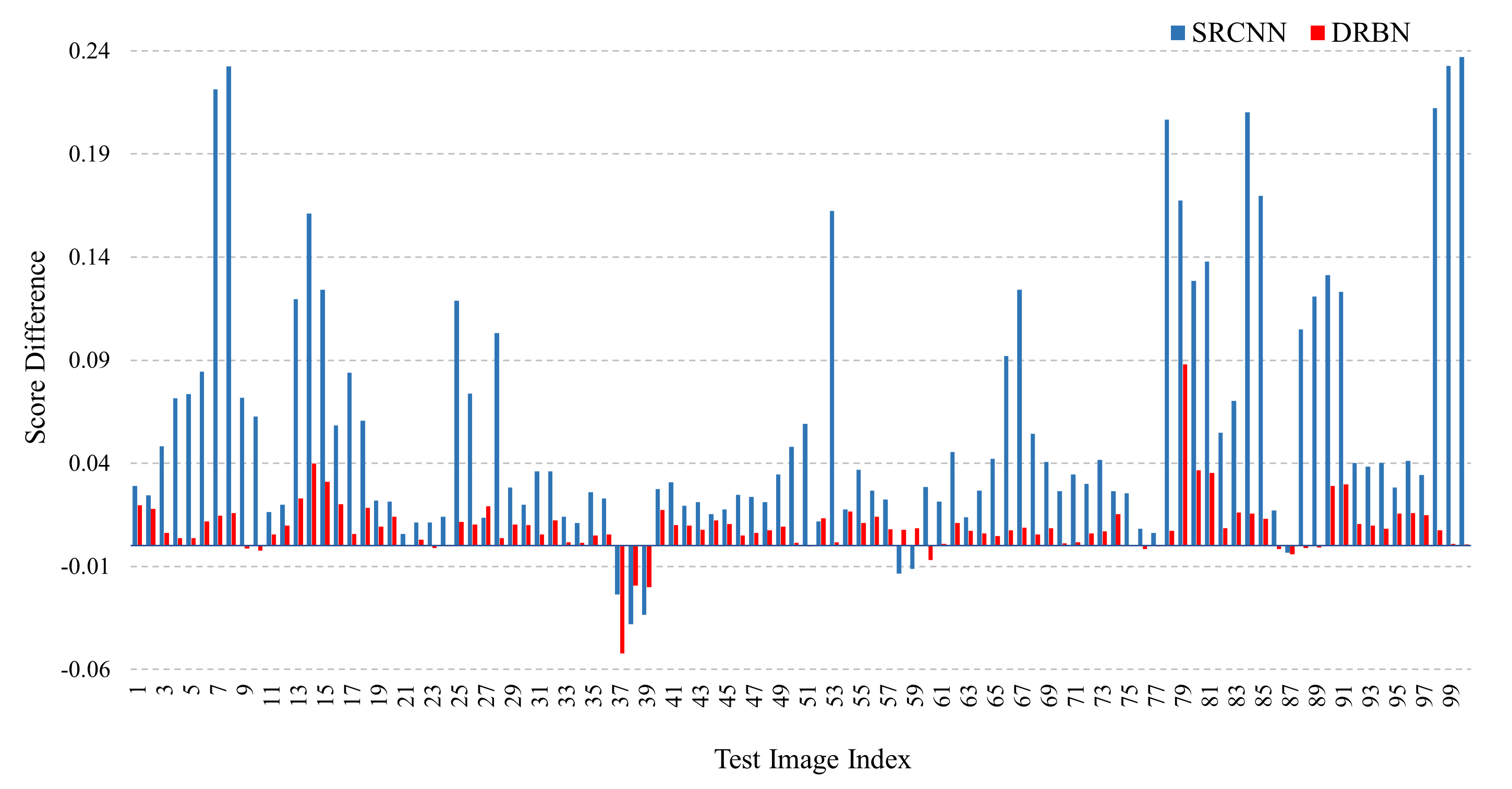}
\caption{\bl{The predicted quality score difference between the optimization results and the original results. 
For each image pair, a positive score difference means the quality of the optimization result is better than the original one. The  SRCNN  and DRBN architectures are utilized for optimization, respectively.}}
%
\label{fig: diffs}
\end{center}
\end{figure*}

\begin{figure*}[htbp]
\begin{center}
\includegraphics[width=1.0\textwidth]{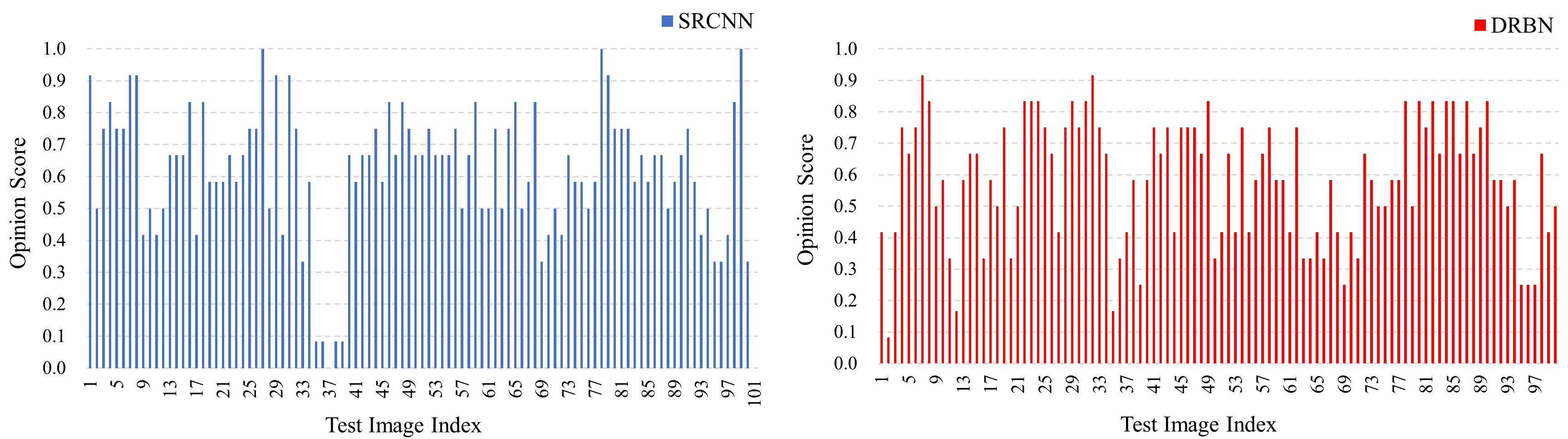}
\caption{The subjective testing results for each image in the testing set are reported. 
The percentage in which the subject is preferred is provided. 
Left: SRCNN; Right: DRBN.} 

\label{fig:opis}
\end{center}
\end{figure*}

\begin{figure*}[h]
\begin{center}
\includegraphics[width=1.0\textwidth]{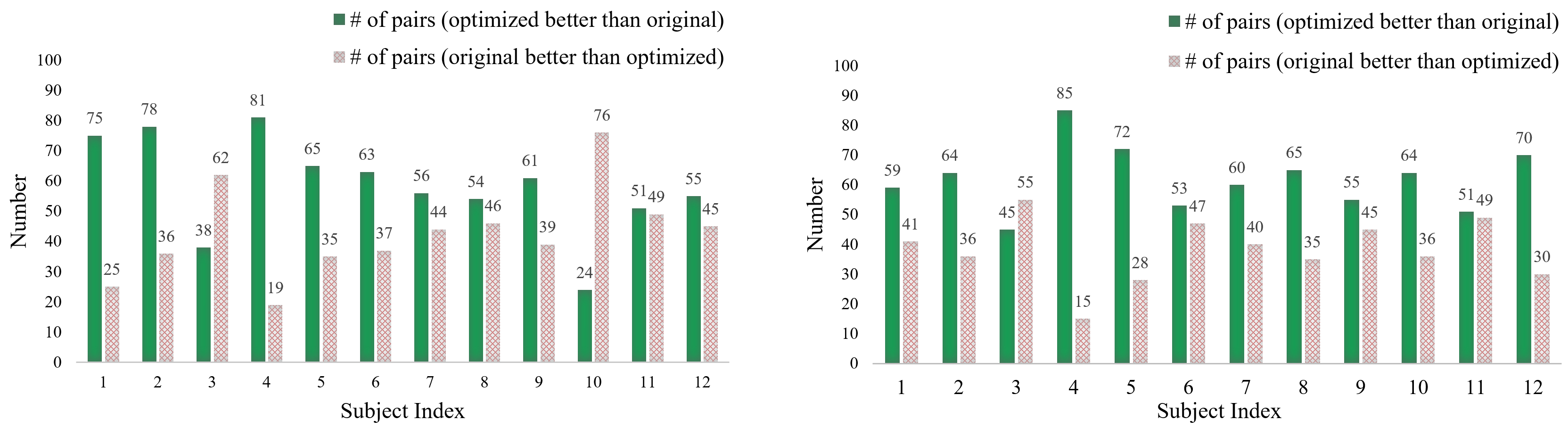}
\caption{The preference of the proposed method and original method for each individual subject on the 100 testing images of LOL~\cite{wei2018deep}. For each image, the two enhanced images by these two methods form a pair for comparison. Left: The SRCNN architecture is adopted for optimization. Right: The DRBN architecture is adopted for optimization.}
\label{fig: SRCNN subjective results}
\end{center}
\end{figure*}

\subsubsection{Ablation Study of the IACA model}
In order to analyze the functionalities of the various modules in our proposed IACA model, we perform the ablation study in this subsection. 
The influence of each design choice is shown in Table~\ref{tab:abl}, and the best result in our experiment is achieved with multi-scale feature extraction, illumination  modulation as well as content adaptation performed. 
More specifically, we first ablate the multi-scale feature learning and use the last stage (Conv5) of ResNeXt for training. 
The results reveal that features in different scales (denoted as MS-Feat in Table \ref{tab:abl}) are more capable of handling such more complicated distortions. 
Subsequently, we ablate the illumination  modulation  from our original model, and significant performance degradation (SROCC from 0.875 to 0.866) can be observed. 
This phenomenon further demonstrates that illumination information plays a critical role in quality perception. 
Moreover, we also analyze different spatial pooling strategies by maintaining the MS-Feat and illumination modulation. 
As shown in Table~\ref{tab:abl}, neither the individual std-pooling nor mean-pooling can achieve the best results. 
By contrast, with our proposed content adaptation module, the content adaptive quality features can be finally acquired, further providing evidence of the effectiveness of our adaptation component.

\subsubsection{Validations on Enhancement Optimization}
In Fig.~\ref{fig:cmpr}, we visually compare the enhanced results of the original baseline as well as the optimized models. 
Our optimization framework delivers visually better results, while the images enhanced by the original SRCNN are darker with lower contrast.
Moreover, the DRBN tends to over-smooth images compared with our method. \bl{In Fig.~\ref{fig: diffs}, we adopt our IACA model for the quality score prediction of the optimization results and the original results, and the score difference of each image pair is presented. From the figure, we can observe that 94\% score difference is positive for SRCNN and 86\% score difference is positive for DRBN models, verifying that most of the optimization results achieve higher quality scores than the original ones.}
This phenomenon demonstrates the effectiveness of the proposed full-chain optimization guided by the perceptual quality of different network architectures.  
%

%

Herein, our target is not to achieve the best enhancement performance. 
The intended message that we are trying to convey is that optimization guided by perceptual characteristics can successfully improve the image quality generated by these widely adopted models. 
%
%
Furthermore, we conduct subjective evaluations by comparing the images generated from original and optimized models. 
We adopt a similar strategy as in the creation of SQUARE-LOL. 
%
During each trial of the experiment, the 2AFC strategy was used to compare the quality of the images generated by the original baseline and the proposed models. 
Specifically, the subject was presented with a pair of images and asked to select the one with better quality. 
In total, 12 subjects participated in this experiment. 
The percentage that the proposed optimization scheme is in favor of against the original model for each image is shown in Fig.~\ref{fig:opis} and the overall percentages are 70 \% for SRCNN and 62 \% for DRBN models, revealing the subjective quality has been improved by a large margin with both SRCNN and DRBN architectures, demonstrating the promising capability of model optimization from a perceptual perspective. 
In Fig.~\ref{fig: SRCNN subjective results}, we also report the preference between the optimized images and the originally enhanced images (without optimization) of each subject. 
We can conclude that more than 90 \% (SRCNN) and 83 \% (DRBN) subjects have more votes for our optimized images when compared with the original ones, further providing evidence that more visually pleasant enhancement results can be obtained by our optimization scheme.

\section{Conclusions}
In this paper, we propose a principled framework for evaluating the subjective quality of low-light images, designing quality assessment algorithms, and optimizing low-light image enhancement methods to improve perceptual quality.
Our experimental findings demonstrate that the proposed objective quality model IACA agrees well with the subjective opinions and reveals the potential in perceptually optimizing the enhancement models in a data-driven fashion. 
While the proposed gap-closing framework represents one of the initial efforts in this area of research, further refinement can be achieved by establishing the sophisticated closed-loop via quality assessment of the enhanced images from the optimized models.
Additionally, exploring the impact of varying image content on quality perception poses novel challenges for IQA and low-light enhancement research, thereby providing impetus for innovative future explorations.
%



\ifCLASSOPTIONcaptionsoff
  \newpage
\fi

\bibliographystyle{IEEEtran}
\bibliography{LLData}

\begin{IEEEbiography}[{\includegraphics[width=1in,height=1.25in,clip,keepaspectratio]{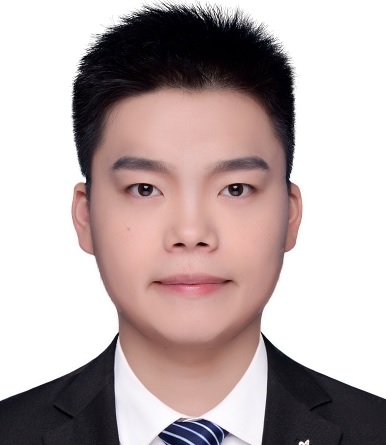}}]{Baoliang Chen} received his B.S. degree in Electronic Information Science and Technology from Hefei University of Technology, Hefei, China, in 2015, his M.S. degree in Intelligent Information Processing from Xidian University, Xian, China, in 2018, and his Ph.D. degree in computer science from the City University of Hong Kong, Hong Kong, in 2022. He is currently a postdoctoral researcher with the Department of Computer Science, City University of Hong Kong.  His research interests include image/video quality assessment and transfer learning.
\end{IEEEbiography}

\begin{IEEEbiography}[{\includegraphics[width=1in,height=1.25in,clip,keepaspectratio]{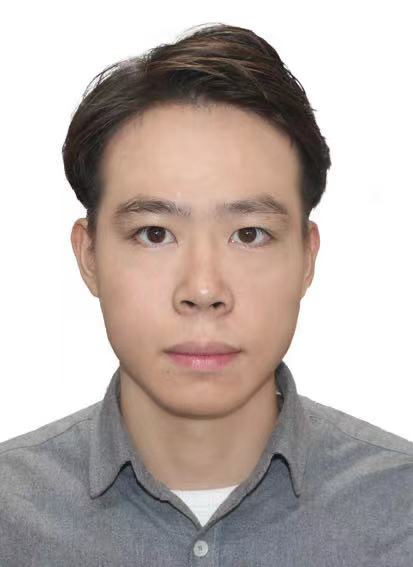}}]{Lingyu Zhu} received the B.S. degree from the Wuhan University of Technology in 2018 and the master's degree from Hong Kong University of Science and Technology in 2019. He is currently pursuing a Ph.D. degree at the City University of Hong Kong. His research interests include image/video quality assessment, image/ video processing, and deep learning.
\end{IEEEbiography}

\begin{IEEEbiography}[{\includegraphics[width=1in,height=1.25in,clip,keepaspectratio]{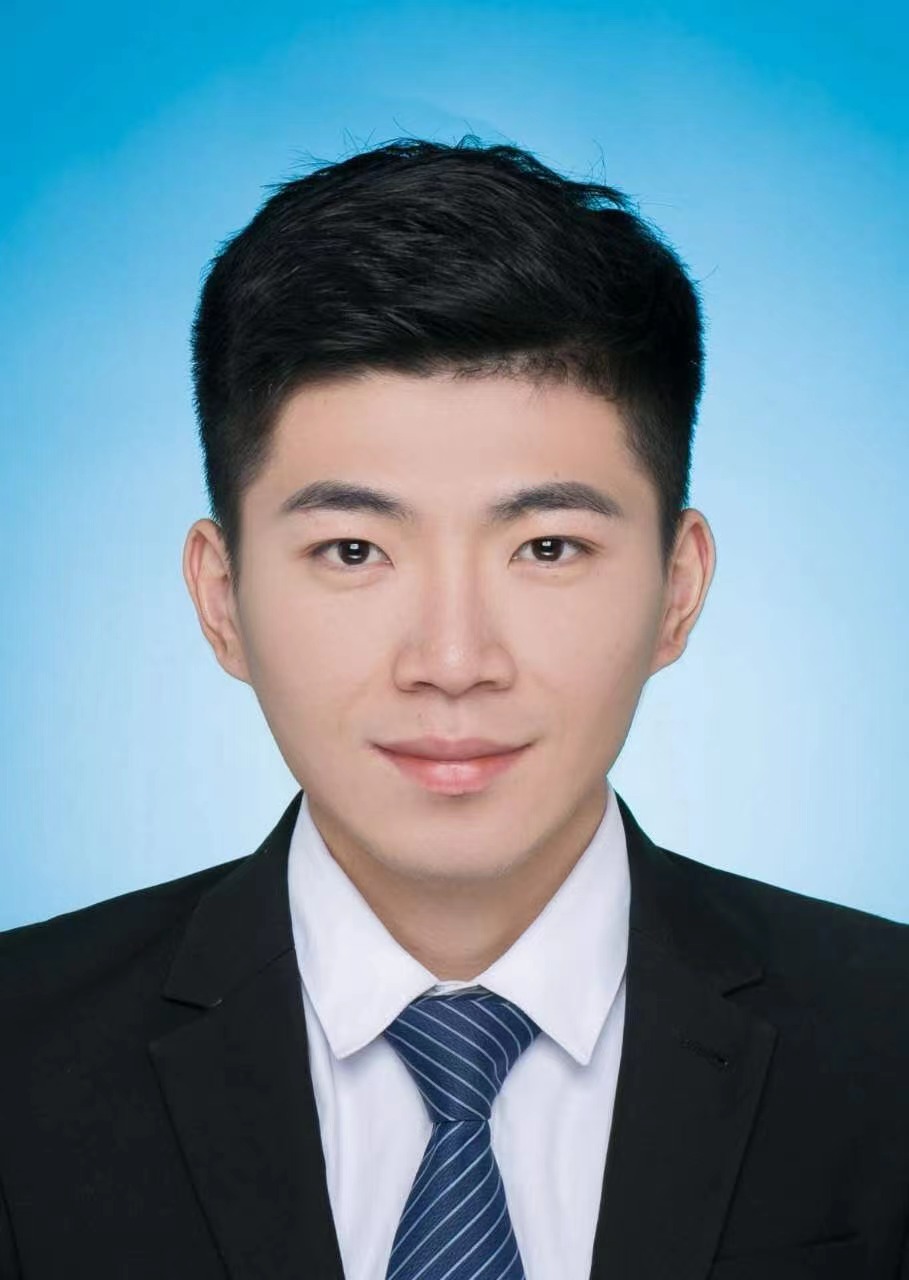}}]{Hanwei Zhu} received the B.E and M.S. degrees from the Jiangxi University of Finance and Economics, Nanchang, China, in 2017 and 2020, respectively. He is currently pursuing a Ph.D. degree in the Department of Computer Science, City University of Hong Kong. His research interest includes perceptual image processing and computational photography.
\end{IEEEbiography}

\begin{IEEEbiography}[{\includegraphics[width=1in,height=1.25in,clip,keepaspectratio]{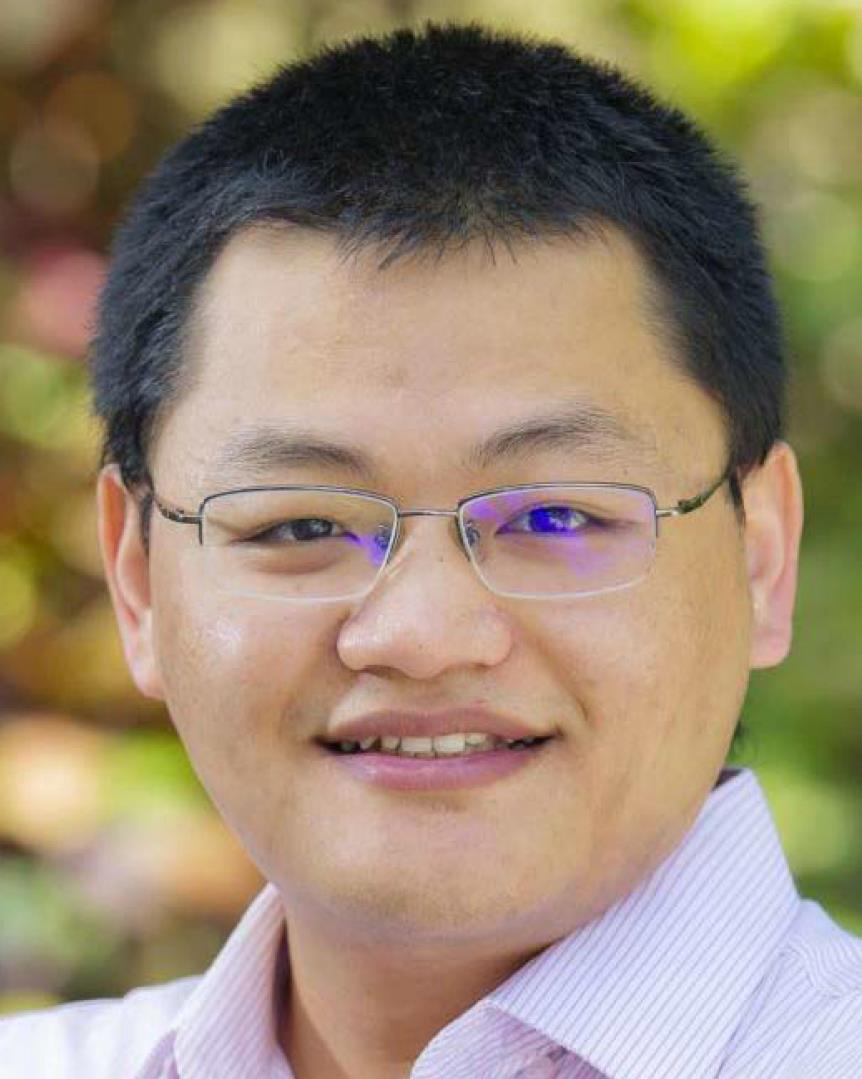}}]{Shiqi Wang} (Senior Member, IEEE) received the B.S. degree in computer science from the Harbin Institute of Technology in 2008 and the Ph.D. degree in computer application technology from Peking University in 2014. From 2014 to 2016, he was a Post-Doctoral Fellow with the Department of Electrical and Computer Engineering, University of Waterloo, Waterloo, ON, Canada. From 2016 to 2017, he was a Research Fellow with the Rapid-Rich Object Search Laboratory, Nanyang Technological University, Singapore. He is currently an Associate Professor with the Department of Computer Science, City University of Hong Kong. He has proposed more than 50 technical proposals to ISO/MPEG, ITU-T, and AVS standards, and authored or coauthored more than 200 refereed journal articles/conference papers. His research interests include video compression, image/video quality assessment, and image/video search and analysis. He received the Best Paper Award from IEEE VCIP 2019, ICME 2019, IEEE Multimedia 2018, and PCM 2017. His coauthored article received the Best Student Paper Award in the IEEE ICIP 2018. He was a recipient of the 2021 IEEE Multimedia Rising Star Award in ICME 2021. He serves as an Associate Editor for \textsc{IEEE Transactions on Circuits and Systems for Video Technology}. 
\end{IEEEbiography}



%









\end{document}